\documentclass[10pt,journal,compsoc]{IEEEtran}

\ifCLASSOPTIONcompsoc
  \usepackage[nocompress]{cite}
\else
  \usepackage{cite}
\fi
\usepackage{graphicx}
\usepackage{amsfonts}
\usepackage{algorithm2e}
\usepackage{subfig}
\usepackage{amsmath}
\usepackage{amssymb}
\usepackage{multirow}
\usepackage{stfloats}
\usepackage{diagbox}
\usepackage{hyperref}
\usepackage{xcolor}
\usepackage{booktabs}
\usepackage{makecell}
\usepackage{caption}
\usepackage{comment}
\usepackage[normalem]{ulem} 
\usepackage{bbding} 
\usepackage{mathtools, nccmath} 
\DeclarePairedDelimiter{\nint}\lfloor\rceil 

\ifCLASSINFOpdf
\else
\fi
\begin{document}
%
\title{Efficient and Generic Point Model for Lossless Point Cloud Attribute Compression}
%
%
%
%

\author{
    Kang You,~
    Pan Gao,~
    and Zhan Ma
            
    \IEEEcompsocitemizethanks{
        \IEEEcompsocthanksitem Kang You and Pan Gao are with the College of Computer Science and Technology, Nanjing University of Aeronautics and Astronautics, Nanjing, China. E-mail: \{youkang, pan.gao\}@nuaa.edu.cn
        \IEEEcompsocthanksitem Zhan Ma is with School of Electronic Science and Engineering, Nanjing University, Nanjing, China. E-mail: mazhan@nju.edu.cn
    }
    
}

%
%

\markboth{Journal of \LaTeX\ Class Files,~Vol.~14, No.~8, August~2015}%
{Shell \MakeLowercase{\textit{et al.}}: Bare Advanced Demo of IEEEtran.cls for IEEE Computer Society Journals}
%




\IEEEtitleabstractindextext{%
\begin{abstract}
The past several years have witnessed the emergence of learned point cloud compression (PCC) techniques. However, current learning-based lossless point cloud attribute compression (PCAC) methods either suffer from high computational complexity or deteriorated compression performance. Moreover, the significant variations in point cloud scale and sparsity encountered in real-world applications make developing an all-in-one neural model a challenging task. In this paper, we propose PoLoPCAC, an efficient and generic lossless PCAC method that achieves high compression efficiency and strong generalizability simultaneously. We formulate lossless PCAC as the task of inferring explicit distributions of attributes from group-wise autoregressive priors. A progressive random grouping strategy is first devised to efficiently resolve the point cloud into groups, and then the attributes of each group are modeled sequentially from accumulated antecedents. A locality-aware attention mechanism is utilized to exploit prior knowledge from context windows in parallel. Since our method directly operates on points, it can naturally avoids distortion caused by voxelization, and can be executed on point clouds with arbitrary scale and density. Experiments show that our method can be instantly deployed once trained on a Synthetic 2k-ShapeNet dataset while enjoying continuous bit-rate reduction over the latest G-PCCv23 on various datasets (ShapeNet, ScanNet, MVUB, 8iVFB). Meanwhile, our method reports shorter coding time than G-PCCv23 on the majority of sequences with a lightweight model size (2.6MB), which is highly attractive for practical applications. Dataset, code and trained model are available at \url{https://github.com/I2-Multimedia-Lab/PoLoPCAC}.
\end{abstract}

\begin{IEEEkeywords}
Point cloud, attribute compression, lossless reconstruction, point model, deep learning.
\end{IEEEkeywords}}

\maketitle

\IEEEdisplaynontitleabstractindextext

%
\IEEEpeerreviewmaketitle

\ifCLASSOPTIONcompsoc
\IEEEraisesectionheading{\section{Introduction}\label{sec:introduction}}
\else
\section{Introduction}
\label{sec:introduction}
\fi

%
%
%
%
\IEEEPARstart{P}{oint} cloud represents 3D objects and scenes through a collection of points, in which each point has a coordinate $(x,y,z)$ and a number of attributes such as color, reflectance, etc. With the fast development of immersive multimedia devices and applications, point cloud has been widely used in numerous tasks such as augmented/mixed reality, robotics, etc \cite{masalkhi2023appleVisionPro, yang2023culturalHeritage, mirzaei20223dPointCloud}. Due to the inherent limitations of networked bandwidth and device storage space, Point Cloud Compression (PCC) is becoming increasingly indispensable to produce compact point cloud representations \cite{valenzise2023PCC}.

\subsection{Background and Motivation}

Owing to the advances of graphics computing platforms and deep learning technologies, recent years have witnessed the emergence of learning-based PCC solutions. Both Moving Picture Experts Group (MPEG) and Joint Photographic Experts Group (JPEG) committees have launched explorations on Artificial Intelligence (AI) based PCC solutions.

In contrary to the significant compression gains demonstrated in learning-based Point Cloud Geometry Compression (PCGC) methods~\cite{wang2022sparsePCGC, fu2022octattention, you2021patch}, learning-based Point Cloud Attribute Compression (PCAC) approaches  are still under development and usually suffering from inferior performance~\cite{nguyen2023MNeT, wang2022sparsePCAC, sheng2021deepPCAC, NFPCAC, RNFPCAC}. Nevertheless, attributes account for the majority of the overall compressed bitstream~\cite{fang20223DAC, fang20224DAC}, especially in lossless mode, which urges the development of effective lossless PCAC techniques.

Current explorations on learning-based lossless PCAC usually extend the multiscale sparse representations originally developed for PCGC~\cite{wang2023losslessPCAC, nguyen2023CNeT, nguyen2023MNeT}. For instance, 3CAC~\cite{wang2023losslessPCAC} is extended from SparsePCGC~\cite{wang2022sparsePCGC} to the attribute compression task. Similarly, a sparse convolution-based pipeline is utilized by CNeT~\cite{nguyen2023CNeT}. An end-to-end multiscale model MNeT~\cite{nguyen2023MNeT} also inherits this framework with optimization for coding efficiency.


Despite the aforementioned efforts, learning-based lossless PCAC still faces the following challenges:

\begin{itemize}
    \item The dilemma of balancing \emph{compression performance} and \emph{computational cost}. These two desiderata present a challenge in terms of their simultaneous fulfillment within one practice.
    
    \item The limitation of \emph{generalizability}. Significant variations in point cloud scale and sparsity encountered in real-world applications make developing an all-in-one neural model a challenging task.
    
    \item Damaged \emph{fidelity} caused by voxelization. The preprocessing of voxelization introduces irreversible distortion that prevent voxel-based models from being truly lossless.
\end{itemize}

In the recent past, point-based models have demonstrated impressive performance in various point cloud processing tasks~\cite{wu2022PTv2,li2023exploiting,PointNN,CompletionSurvey,TVCG_DPCA} with affordable complexity and high flexibility, motivating us to look towards a point model-driven lossless PCAC solution that offers \emph{superior coding efficiency}, \emph{robust generalizability}, and \emph{high fidelity} simultaneously. 

\subsection{Our Approach}
This paper proposes a point model-based solution, dubbed PoLoPCAC, to meet the above-mentioned challenges. Unlike prevalent learning-based lossless PCAC methods, which are constrained to voxelized Point Cloud Geometry (PCG) for processing, we propose to compress point cloud attributes on raw PCG anchors.


We formulate lossless PCAC as a task of inferring explicit distributions of attributes from group-wise autoregressive priors. Inspired by the Level of Detail (LoD) representation~\cite{graziosi2020overview} that re-organizes the input point cloud into refinement levels, we devise a Progressive Random Grouping (PRG) strategy to efficiently resolve the point cloud into mutually exclusive groups. Instead of using the iterative Euclidean distance calculation for LoD generation, we efficiently divide the point cloud into groups using shuffling and splitting operations, resulting in a convenient partition with \emph{linear complexity}. Based on that, the point cloud attributes are modeled in an \emph{inter-group autoregressive} and \emph{intra-group parallel} way, i.e., each group are coded sequentially by considering accumulated antecedent groups as prior knowledge, and the attributes of the points in the same group are coded in parallel. Moreover, this group-wise modeling scheme offers a wide range of scalability levels~\cite{JPEGPlenoCFP}, i.e., it is possible to first decode a coarse point cloud and then progressively refine it with additional bit streams.



Further, adaptive-size context windows are built based on $K$ Nearest Neighbor ($K$NN) graphs, followed by neural network-powered inference, to estimate the distributions of the target attributes. Previous sparse convolution~\cite{SparseConv} based methods~\cite{wang2023losslessPCAC,nguyen2023CNeT,nguyen2023MNeT} mostly use fixed-size convolution kernels (e.g., $3^3$ or $5^3$) to extract features on Positively Occupied Voxels (POV), but this scheme inherently suffers from restricted receptive field that struggles to capture sufficient neighborhoods, especially on relatively sparse PCG surfaces. Instead, we perform neighborhood gathering by building $K$NN graphs on raw points, to attain adaptive-size context windows, regardless of the sparsity of the PCG anchor. In addition, unlike other methods that repetitively use $K$NN query for multiscale aggregation, we only use it to obtain a small context window for each point once, which maintains low coding complexity.

During the neural network inference, Spatial Normalization (SN) operation is first devised to facilitate generic compression for arbitrary point cloud scale and density, by aligning the obtained context windows to a standard spatial volume. Then, normalized windows are fed into the attention-based neural network, producing the parameters for the estimated distribution of the target attributes, to further conduct efficient arithmetic coding in parallel within each group.

\subsection{Contribution}

Main contributions of this work can be summarized as:
\begin{itemize}
    \item To the best of our knowledge, this is \emph{the first} exploration of point model motivated lossless PCAC. By utilizing efficient inter-group autoregressive modeling and intra-group parallel coding, our model achieves superior compression performance.
    
    \item The proposed PoLoPCAC demonstrates significantly low complexity and a portable model size. It reports shorter coding time than the lastest G-PCCv23 on the majority of sequences with on a comparatively low-end platform (e.g., one RTX 2080Ti GPU), which is beneficial for industrial applications.
    
    \item Our model generalizes very well across a variety of test sets once trained on a small synthetic dataset. The point-based pipeline can be executed on point clouds of arbitrary scale and density, without requiring any prior knowledge of the test domain, and providing scale-invariant coding efficiency.

    \item Since our model  operates directly on the points, it inherently avoids distortion caused by voxelization, thus achieving high fidelity.

\end{itemize}

In addition, current learning-based PCAC still faces the problem of inadequate training data. To address this issue, we build a colorized point cloud dataset Synthetic 2k-ShapeNet to facilitate generic training.

\section{Related Work}

In real-world applications, both lossy and lossless PCAC have garnered significant attention, albeit with distinct emphases. This section presents a brief review of relevent lossy\&lossless PCAC works.

\subsection{Lossy Point Cloud Attribute Compression} 

Lossy PCAC aims to minimize data size by allowing for an acceptable level of quality loss during the coding process. To this end, rules-based PCAC methods proficiently capitalize on the statistical redundancies that exist within the transformed frequency domains. The Region-Adaptive Hierarchical Transform (RAHT) \cite{de2016RAHT}, which resembles an adaptive variation of Haar wavelet transform, is further adopted in the MPEG G-PCC standard for its high efficiency. Subsequently, RAHT is extended with plenty of model-driven \cite{sandri2019RAHTinteger, pavez2021RAHTmulti} solutions to further attain superior performance. Other transformations, such as Graph Fourier Transform (GFT), are also widely examined in diverse explorations \cite{xu2020GFTpredictive, song2022GFTfine, song2022GFTrate}. 

Following the recent advancement of learned image compression technologies~\cite{balle2016end2end,balle2018variational,he2021checkerboard,QARV}, data-driven lossy PCAC methods have been emerged. For instance, 3DAC~\cite{fang20223DAC} propose a deep entropy model that efficiently exploits the correlation of RAHT coefficients. As a improved model for dynamic PCAC, 4DAC~\cite{fang20224DAC} build the 3D motion estimation and motion compensation modules to reduce inter-frame redundancy. LVAC~\cite{LVAC} combines the ideas of RAHT and implicit function, yet with an unacceptable encoding time is reported for the iterative fitting process. Some works like DeepPCAC \cite{sheng2021deepPCAC}, SparsePCAC \cite{wang2022sparsePCAC}, NF-PCAC~\cite{NFPCAC}, and RNF-PCAC~\cite{RNFPCAC} are proposed to compress attributes under novel end-to-end autoencoder structures, but their performances are still inferior to the latest G-PCC, especially at relatively high bitrates.

\subsection{Lossless Point Cloud Attribute Compression}

Lossless PCAC reduces data sizes without sacrificing any information in the original point cloud attributes. As a benchmark, G-PCC still dominates on lossless PCAC, on account of the Predicting/Lifting Transform that relies on the Level of Detail (LoD) representation \cite{graziosi2020overview}. Some works improve G-PCC with hand-crafted schemes such as normal-based prediction \cite{yin2021losslessNIP} or content-adaptive LoD \cite{wei2022losslessLOD}, but only with indistinctive gains. 

Learning-based lossless PCAC methods, on the other hand, often follow the inertial thinking of directly inheriting ready-made frameworks from geometry compression. For example, recent work 3CAC~\cite{wang2023losslessPCAC} that extends the multiscale structure from SparsePCGC \cite{wang2022sparsePCGC} can achieve certain gains for several datasets, but suffers from inefficient coding speed and limited generalization capability. CNeT \cite{nguyen2023CNeT} utilizes sparse tensor-based deep neural networks to learn the conditional probability of attributes. However, they achieve state-of-the-art performance by fitting on a limited sequence of data using a massive model with hundreds of millions of parameters. MNeT \cite{nguyen2023MNeT} is a recently proposed method with relatively small model size and affordable complexity but at the expense of compression rates. Moreover, the above-mentioned works need manual pre-processing steps, i.e., re-scale and voxelization, since they are unable to match point clouds with arbitrary scale and density as flexibly as the point model.


\section{Our Approach}

\subsection{Overview}

\begin{figure*}[t]
    \centering
    \includegraphics[width=1\textwidth]{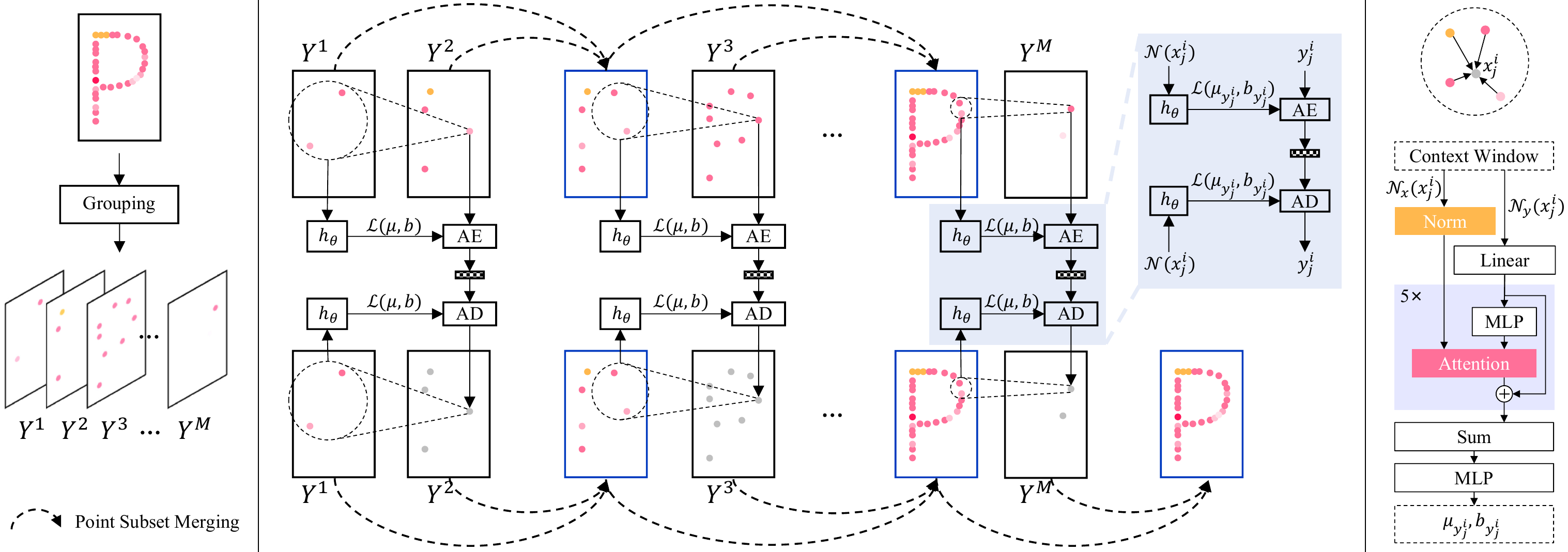}
    \caption{{\bf PoLoPCAC}. Input point cloud is first resolved into several groups (left); Then, attributes of each group are predicted sequentially using accumulated antecedents as context (middle); A locality-aware attention mechanism is utilized to infer the probability distribution of attributes from context windows in parallel (right).}
    \label{fig:framework}
\end{figure*}

This paper mainly focuses on the lossless PCAC task under the premise of a separately and losslessly transmitted geometry. As shown in Fig.~\ref{fig:framework}, we propose an efficient lossless PCAC method by naturally inferring explicit distributions of attributes from group-wise autoregressive priors. Intuitively, points within a miniature scope are semantically related and may share similar attribute values, and the relation increases as the scope shrinks. Inspired by this, attributes are compressed in an inter-group autoregressive and intra-group parallel way. The decoding process is in accordance with the encoding.


\subsection{Problem Definition}

Let $X=\left \{ x_1,x_2,\dots,x_N  \right \}$ represent the set of positions associated with the points of the input point cloud, and $Y=\left \{ y_1,y_2,\dots,y_N \right \}$ be the corresponding attributes, in which $N$ refers to the number of points in the point cloud. Our goal is to perform lossless compression of $Y$ by considering $X$ as auxiliary information.

To achieve this, a model parameterized by $\theta$ is constructed to formulate a parametric probability distribution $P_{\theta }(Y)$, for the approximation of the actual distribution $P(Y)$. Consequently, the optimization for bitrate can be described as:
\begin{equation}
    {\theta}^{\ast} \leftarrow \underset{\theta}{argmin} \mathbb{E}_{P(Y)} \left [ -\log_{2}{P_{\theta }(Y)}  \right ] 
    \label{eq:optimization}
\end{equation}


\subsection{Progressive Random Grouping}
\label{sec:prg}

Inspired by the Level of Detail (LoD) representation that re-organizes the input point cloud into refinement levels \cite{graziosi2020overview}, we propose a Progressive Random Grouping (PRG) strategy to efficiently resolve the input point cloud into groups. In brief, a grouping function $g$ is devised to split the input point cloud into $M$ mutually exclusive groups, given the size of each group $\left \{ s^{i} \right \}_{i=1}^{M}$:
\begin{equation}
    \left \{ X^i, Y^i \right \}_{i=1}^{M}=g(X,Y,\left \{ s^{i} \right \}_{i=1}^{M} )
\end{equation}
\begin{equation}
    X^i=\left \{ x_{1}^{i}, \dots, x_{s^i}^{i} \right \}, Y^i=\left \{ y_{1}^{i}, \dots, y_{s^i}^{i} \right \} 
\end{equation}
To be specific, we first sample $s^1$ points randomly from the original point set as our first group, i.e., $X^1$ and $Y^1$, and then sample $s^2$ points randomly from the leftovers as the second group $X^2$ and $Y^2$. Analogously, the $i$th sampling will be conducted on $N-\sum_{j<i} s^j$ points to generate $s^i$ points as our $i$th group $X^i$ and $Y^i$.

We increase the size of each group gradually to further model the context in a coarse-to-fine manner. The size of $i$th group $s^i$ is defined as follows:
\begin{equation}
    s^1 = min\left \{ \left \lfloor \frac{N}{\alpha}  \right \rfloor,s^{\ast} \right \} 
\end{equation}
\begin{equation}
    s^i=min\left \{ r \times s^{i-1},N-\sum_{j<i}{s^j},s^{\ast}  \right \} 
\end{equation}
where $s^1$ represents the size of the very first group, which is predefined by a fixed ratio of $N$. The size of $i$th group $s^i$ will be expanded $r$ times with respect to the size of previous group $s^{i-1}$. An upper limit $s^{\ast}$ serves as the control of granularity as well as a flexible limitation for GPU memory usage. We set $r$=2, $\alpha$=128 and $s^{\ast}$=2\textsuperscript{14} in our experiment.

\begin{figure}[t]
    \centering
    \includegraphics[width=1.0\linewidth]{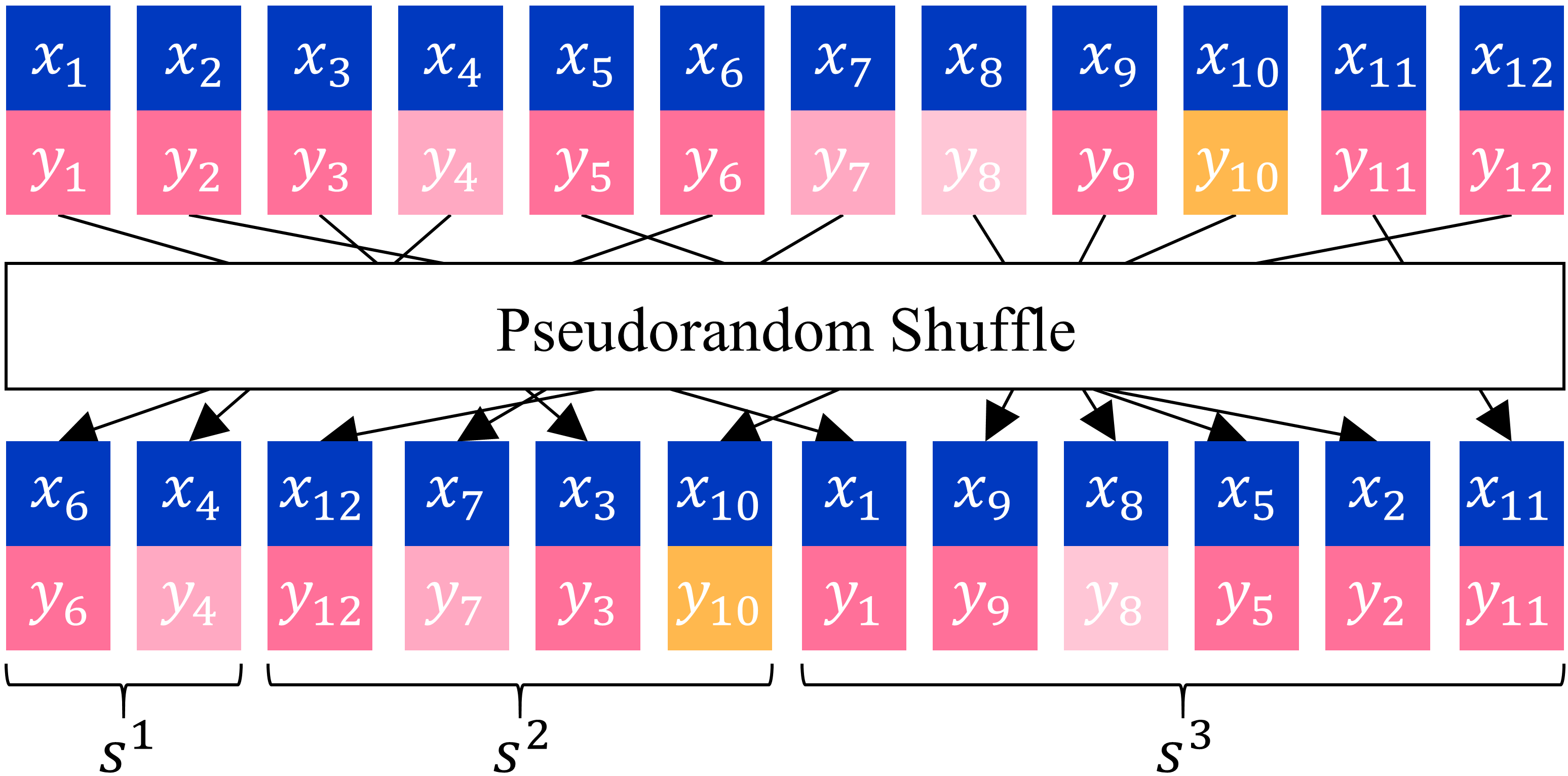}
    \caption{An illustration of fast grouping strategy for practical implementation.}
    \label{fig:grouping_implementation}
\end{figure}

For implementation, we merge multiple random point sampling as a single shuffle function and achieve the grouping process by a split operation. A simple sketch is shown in Fig.~\ref{fig:grouping_implementation} for better understanding. It should be noted that our shuffle function is a pseudorandom process with a fixed seed shared between encoder and decoder, in order to ensure consistent grouping results.

\subsection{Group-Wise Autoregressive Modeling}
\begin{figure}[t]
    \centering
    \includegraphics[width=0.7\linewidth]{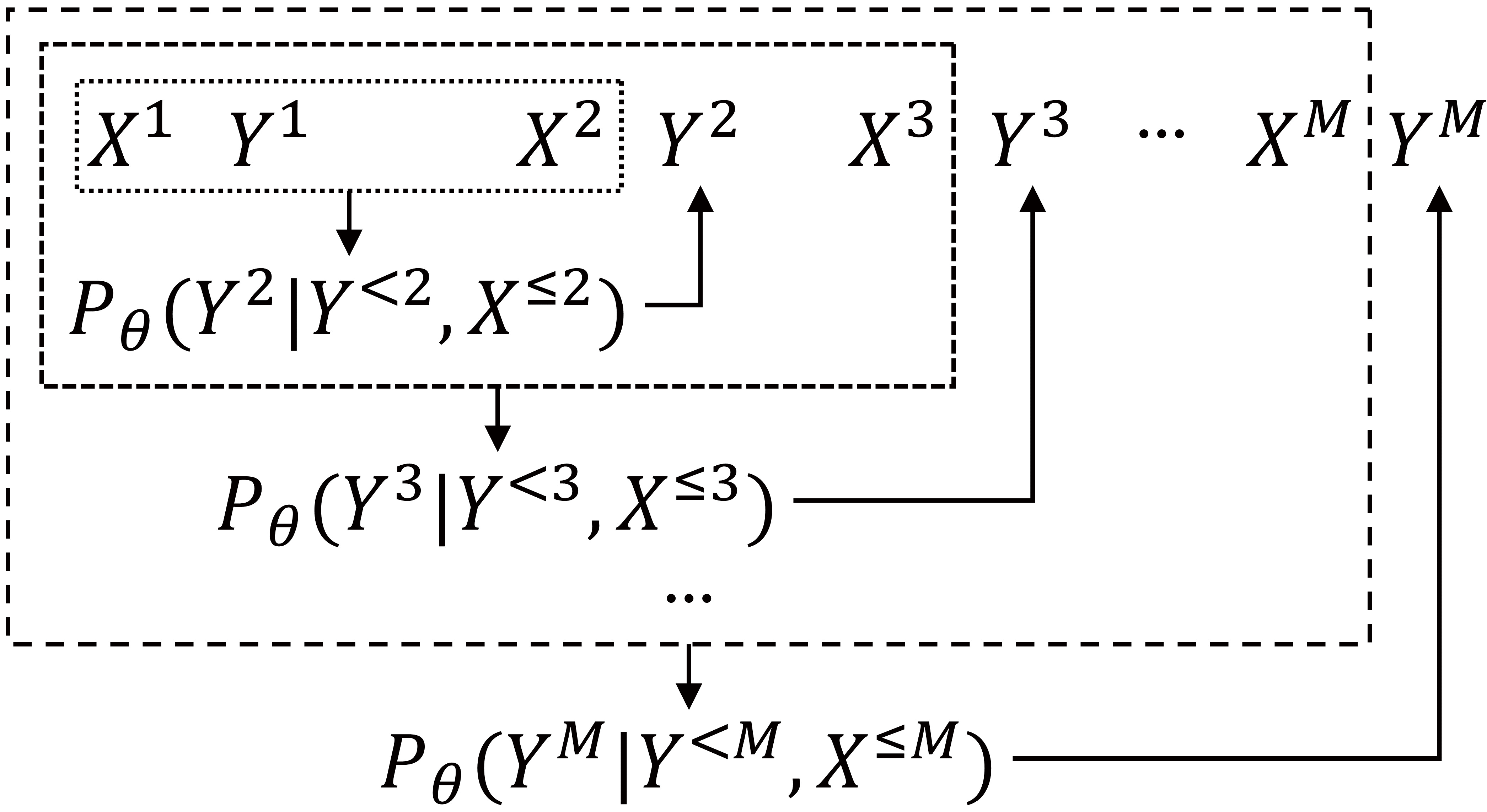}
    \caption{Operational diagram for our group-wize autoregressive modeling. Each group is modeled serially by using previous groups as context.}
    \label{fig:op_diagram}
\end{figure}

With a series of randomly divided groups, we model our lossless PCAC task using a group-wize factorization. Here, the geometry $X$ is introduced as auxiliary information to aid the group-wise autoregressive model:
\begin{align}
\begin{split}
    P_{\theta}(Y) & = \frac{P_{\theta}(Y|X)  \overbrace{P_{\theta}(X)}^{=1}}{\underbrace{P_{\theta}(X|Y)}_{=1}}\\
                    & = P_{\theta}(Y|X) \\
                    & = P_{\theta}(Y^1,Y^2,\dots,Y^M|X) \\
                    & = P_{\theta}(Y^1|X^1)\prod_{i=2}^{M} P_{\theta}(Y^i|Y^{<i},X^{\le i})
\end{split}
\end{align}
In this instance, the distribution of attribute $Y^i$ corresponding to the $i$th group is analyzed by a conditional probability with respect to previous groups ($Y^{<i}$, $X^{<i}$) and $X^i$ that represents the corresponding positions of the attribute $Y^i$. An operational diagram is shown in Fig.~\ref{fig:op_diagram} for a better understanding. It should be noted that the attributes within the first group, i.e., $Y^1$, are directly encoded under uniform distribution.

By using information from previous groups as context, intra-group points can be further modeled through an efficient parallel inference, i.e., 
\begin{align}
\begin{split}
  P_{\theta}(Y^i|Y^{<i},X^{\le i}) & = \prod_{j=1}^{s^i} P_{\theta}(y_{j}^i|x_{j}^i,Y^{<i},X^{<i}) \\
                            & = \prod_{j=1}^{s^i} \int_{y_j^i-\frac{1}{2}}^{y_j^i+\frac{1}{2}}  \mathcal{L}(y|{\Phi}_{y_j^i}) \; d y
\end{split}
\label{eq:intra_group_factorize}
\end{align}
where $\mathcal{L}({\Phi}_{y_j^i})$ represents the probability density function (PDF) of Laplace distribution of attribute $y_j^i$ with estimated location and scale parameters ${\Phi}_{y_j^i} = ({\mu }_{y_{j}^{i}},{b}_{y_{j}^{i}})$. Furtherly, parameter ${\Phi}_{y_j^i}$ is predicted by a parameter inference network $h_{\theta}$ that takes a local context window $\mathcal{N}(x^i_j)$ as input: 
\begin{equation}
    {\Phi}_{y_j^i} = ({\mu }_{y_{j}^{i}},{b}_{y_{j}^{i}}) = h_{\theta }(\mathcal{N}(x^i_j))
\end{equation}
where $\mathcal{N}(x^i_j)$ is formulated as a neighboring scope drawn by a $K$-nearest neighbor function $F_{knn}$ based on Euclidean distance, with respect to the target position $x^i_j$ and accumulated antecedents $\sum X^{<i}$: 
\begin{equation}
    \mathcal{N}(x^i_j) = \left\{ (x_k, y_k) | k \in F_{knn}(x^i_j, \sum X^{<i}) \right\}
\end{equation}

\subsection{Network}

As mentioned in the previous subsection, we use a neural network $h_{\theta}$ to infer distribution parameter ${\Phi}_{y_j^i}$ from the local context window $\mathcal{N}(x^i_j)$. The structure of our network is shown in the right part of Fig.~\ref{fig:framework}, where $\mathcal{N}_{x}(x^i_j)$ represents the point positions, and $\mathcal{N}_{y}(x^i_j)$ represents the corresponding attributes.

\subsubsection{Spatial Normalization}
As the scale and density of point cloud data vary considerably, we first use a Spatial Normalization (SN) operation to eliminate this disparity, as well as align the context window to the target position $x^i_j$. Specifically, each coordinate within the window is normalized as below:
\begin{equation}
\frac{x_k - x_j^i}{\underset{{x}_k' \in \mathcal{N}_x(x_j^i) }{max} \left \{ \left \| {x}_k' - x_j^i \right \|_2  \right \}  } ,\quad x_k \in \mathcal{N}_x(x_j^i)
\label{eq:spatial_norm}
\end{equation}
Meanwhile, the corresponding attributes are mapped into a higher dimension by a linear projection ${\psi}_{map}$. Denote the normalized coordinates as $\mathcal{P}$ and mapped attributes as $\mathcal{F}^{(1)}$, the above-mentioned process can be represented as follows: 
\begin{equation}
    \mathcal{P} = Norm(\mathcal{N}_{x}(x_j^i)), \quad \mathcal{F}^{(1)} = {\psi}_{map}(\mathcal{N}_{y}(x_j^i))
\end{equation}
Let $K$ be the number of points in the window, then $\mathcal{P}$ and $\mathcal{F}^{(1)}$ are in the shape of $K \times 3$ and $K \times C$, respectively.

\subsubsection{Query-Masked Attention}
An attention mechanism is applied to aggregate $\mathcal{P}$ and $\mathcal{F}^{(1)}$ with skip connections, where an input embedding ${\delta}_{emb}$ is first utilized to increase the non-linearity of our network: 
\begin{align}
\begin{split}
\mathcal{F}^{(l+1)} & = \bigtriangleup \mathcal{F}^{(l)} + \mathcal{F}^{(l)} \\
                  & = Attn^{(l)}(\mathcal{P},\mathcal{F}^{(l)}_{emb}) + \mathcal{F}^{(l)} \\
                  & = Attn^{(l)}(\mathcal{P},{\delta}_{emb}(\mathcal{F}^{(l)})) + \mathcal{F}^{(l)} \\
\end{split}
\end{align}
where $\mathcal{F}^{(l)}$ means the identity feature for the $l$th residual unit, $\bigtriangleup \mathcal{F}^{(l)}$ means the output additive feature of $l$th attention module $Attn^{(l)}$, and $\mathcal{F}_{emb}^{(l)}$ represents the embedded input feature for our attention module.

\begin{figure}
    \centering
    \includegraphics[width=1.0\linewidth]{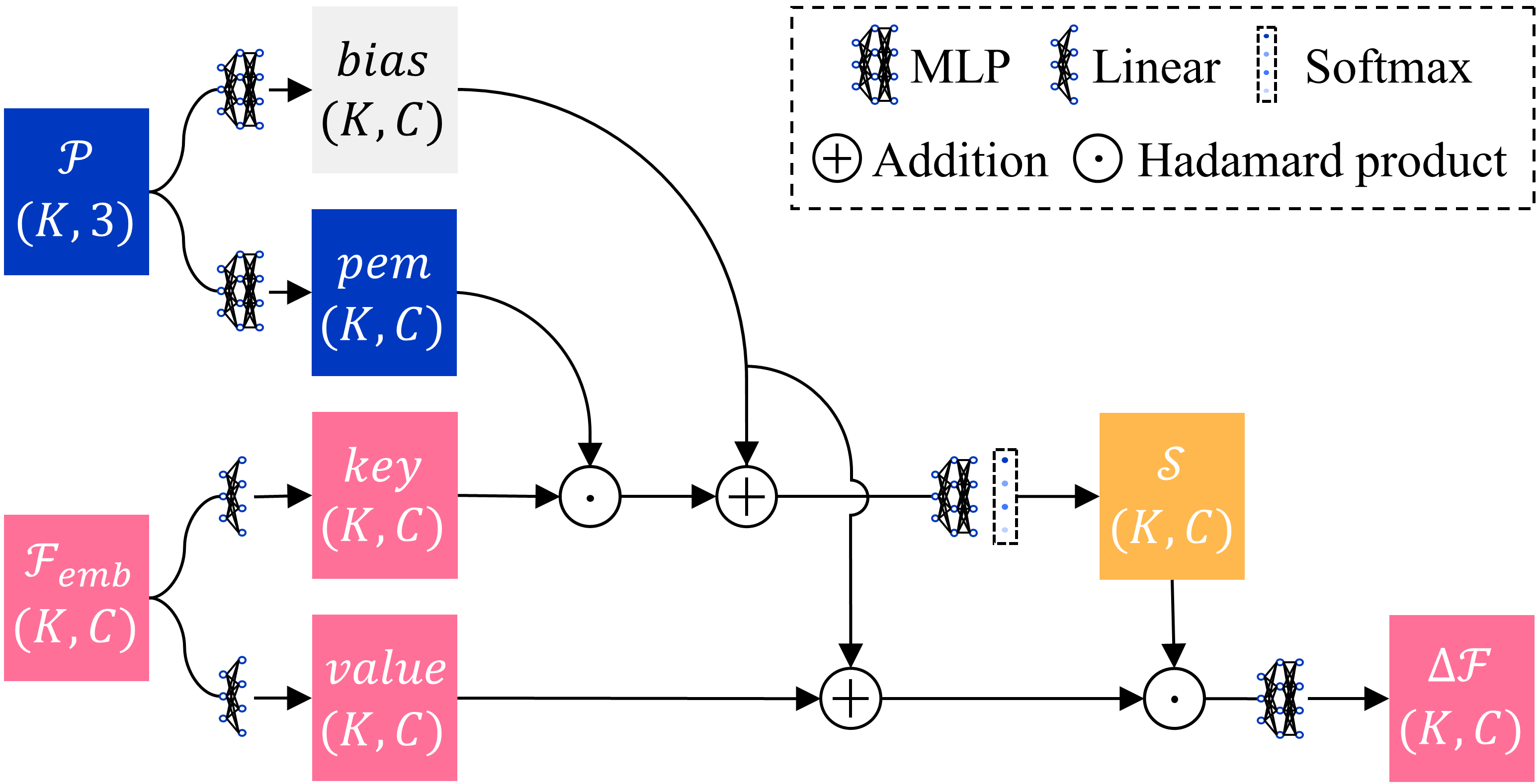}
    \caption{Attention mechanism used in this work. \emph{query} of the target point is masked to zero thus \emph{key} is directly utilized to fuse positional encodings. \emph{pem} refers to the position encoding multiplier. Superscript is omitted for a concise explanation.}
    \label{fig:attention}
\end{figure}

Conventional attention usually project point features to the feature vectors of \emph{query}, \emph{key}, and \emph{value}, then the \emph{key} of the neighbor points is subtracted from the \emph{query} of the target point as the subtraction relation \cite{zhao2021PTv1, wu2022PTv2}. However, in our task, the feature of the target point is not available, since distribution of the target attribute is exactly what we need to infer. Therefore, we mask the \emph{query} to zero, thus utilizing \emph{key} to directly fuse positional encodings.

The detail of our used attention mechanism is illustrated in Fig.~\ref{fig:attention}. The position encoding is considered as \emph{pem} (position encoding multiplier) and \emph{bias} to obtain the spatial information, inspired by PTv2 \cite{wu2022PTv2}. Mathematically,
\begin{equation}
\mathcal{S}^{(l)} =\sigma ( {\delta}_{pem}^{(l)}(\mathcal{P}) \odot {\psi}_{key}^{(l)}(\mathcal{F}^{(l)}_{emb}) + {\delta}_{bias}^{(l)}(\mathcal{P}))
\label{eq:attn_calc_score}
\end{equation}
\begin{equation}
\bigtriangleup \mathcal{F}^{(l)} = {\delta}^{(l)}_{out}(({\psi}^{(l)}_{value}(\mathcal{F}^{(l)}_{emb})+{\delta}^{(l)}_{bias}(\mathcal{P}))\odot \mathcal{S}^{(l)})
\label{eq:attn_calc_fea}
\end{equation}
where ${\delta}^{(l)}$ and ${\psi}^{(l)}$ represent multilayer perceptrons (MLPs) and linear layers in $Attn^{(l)}$, respectively. $\odot$ refers to the Hadamard product and $\sigma$ refers to the Softmax function that acts point-wisely within a context window.

In this fashion, feature $\mathcal{F}^{(L)}$ with a shape of $K \times C$ is produced by $L$th unit. A $Sum$ function is utilized to aggregate $\mathcal{F}^{(L)}$ to a $1 \times C$ feature, and then the parameter ${\Phi}_{y^i_j}$ is predicted through a multilayer perceptron ${\delta}_{head}$:
\begin{equation}
    {\Phi}_{y^i_j}={\delta}_{head}(Sum(\mathcal{F}^{(L)}))
\end{equation}
Consequently, the probability distribution of $y^i_j$ is formulated by the predicted parameter ${\Phi}_{y^i_j}$, as mentioned in Eq.~\ref{eq:intra_group_factorize}, to further code attribute signals.

We set $K=8$, $L=5$ and $C=128$ in our experiment.

\subsubsection{Loss Function}
Finally, since we train our network by minimizing the difference between actual distribution $P(Y)$ and predicted distribution $P_{\theta}(Y)$ in a group manner, a group-wise cross-entropy loss is considered as our loss function:
\begin{equation}
    Loss = \frac{1}{M} \sum_{i=1}^{M} H(P(Y^i), P_{\theta}(Y^i))
\end{equation}

\section{Experiments}
\label{sec:experiments}


\subsection{Experimental Setup}
\label{sec:exp_setup}

\begin{figure*}[t]
    \centering
    \includegraphics[width=1\textwidth]{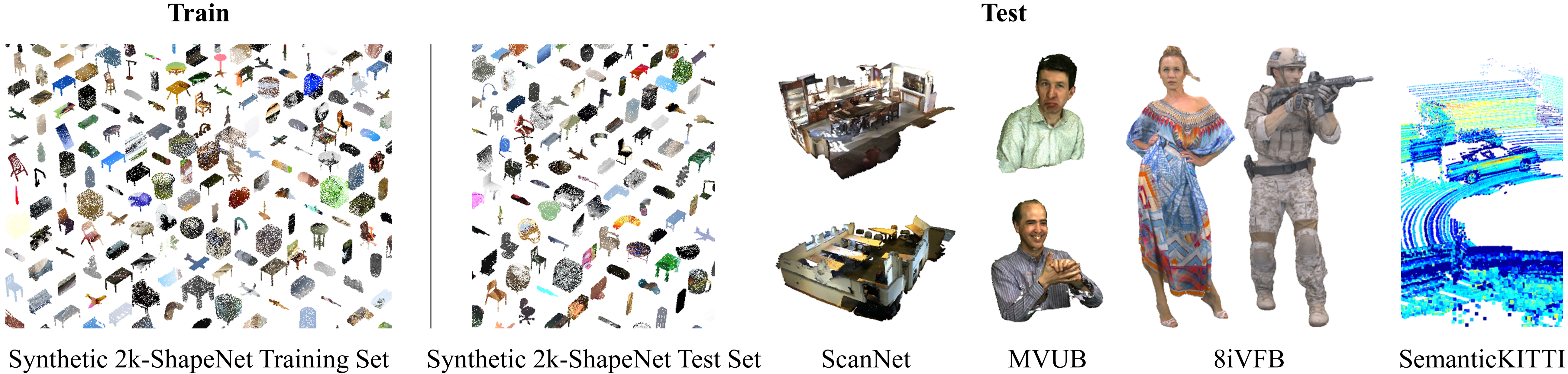}
    \caption{Datasets used in this paper. We limit the training on the Synthetic 2k-ShapeNet training set and test on a variety of datasets.}
    \label{fig:dataset_overview}
\end{figure*}

\subsubsection{Dataset}

As shown in Fig.~\ref{fig:dataset_overview}, we limit the training on our Synthetic 2k-ShapeNet training split only, and then test it on a variety of datasets.

\begin{itemize}
    \item \textbf{Synthetic 2k-ShapeNet}. Due to the lack of a widely-used training set in PCAC \cite{nguyen2023MNeT}, we bring a colorized sparse point cloud dataset to the community. Inspired by SparsePCAC \cite{wang2022sparsePCAC}, we first sample 2048 (2k) points from raw ShapeNet \cite{chang2015shapenet} meshes, then colorize each point cloud by an image that is randomly selected from PCCD \cite{chang2017PCCD}. To be specific, for each sampled point cloud, we project the points to a 2D pixel grid by a Polar-to-Euclidean coordinate transform, then each point will be colored based on the image pixels corresponding to the grid. Following the official training/testing split, point clouds are split into 35,708 for training, 5,158 for validation, and 10,261 for testing. Examples can be found in Fig.~\ref{fig:dataset_overview}. Please refer to our supplementary appendix for more details.

    Since the methods we compared do not support the input of point cloud within unit space, here, we rescaled the coordinates of each point cloud to the range of [0, 1023], abbreviated as 10bit, to minimize the error caused by voxelization as much as possible. Meanwhile, we also conduct evaluation on the range of [0, 63], a.k.a., 6bit, to present comprehensive results at different scales.
    
    \item \textbf{ScanNet}. As one of the most representative point cloud datasets for indoor scenes, ScanNet \cite{dai2017scannet} has 1513 scans generated by RGB-D cameras. We use 312 scans for test as suggested, with about 30k $\sim$ 440k points in each scan. Experiments are examined on the scales of 10bit and 12bit.

    \item \textbf{MVUB \& 8iVFB}. Microsoft Voxelized Upper Bodies (MVUB) \cite{loop2016MVUB} and 8i Voxelized Full Bodies \cite{d20178iVFB} are two widely used human body datasets in standardization committee. Unlike ShapeNet and ScanNet, point clouds in MVUB\&8iVFB are born with the points being dense and voxelized  at the first stage. We use all sequences for evaluation.

    \item \textbf{SemanticKITTI}. SemanticKITTI \cite{behley2019SemanticKITTI} is a large-scale LiDAR point cloud dataset with a total of 43,552 scans available. We use sequence 11, \emph{i.e.}, the first sequence in suggested test set to examine our compression efficiency on point cloud reflectance.
\end{itemize}

We classify the above-mentioned datasets into three categories, i.e., \emph{Object\&Scene} including Synthetic 2k-ShapeNet and ScanNet, \emph{HumanBodies} including MVUB and 8iVFB, and \emph{LiDAR} including SemanticKITTI sequence 11.


\subsubsection{Implementation}

We implement our model using Python 3.10 and Pytorch 2.0. Adam optimizer \cite{kingma2014adamOpt} is used with an initial learning rate of 0.0005 and a batch size of 8. We train our model with 170,000 steps in our Synthetic 2k-ShapeNet training set. The YCoCg-R~\cite{malvar2003ycocg} color space is used in this paper. All experiments are performed on Intel Core i9-9900K CPU and one NVIDIA RTX2080TI GPU.

\subsubsection{Benchmarking Baseline}






We compare our method with the state-of-the-art methods, including rules-based model G-PCCv23~\cite{GPCCv23} and learning-based models CNeT~\cite{nguyen2023CNeT} and MNeT~\cite{nguyen2023MNeT}. For fair comparison, we retrain CNeT and MNeT under our 6bit Synthetic 2k-ShapeNet training split, then test these models under the same condition as our model. It takes about two weeks for CNeT and several hours for MNeT to converge. Note that the YCoCg-R and RGB color space are utilized in CNeT and MNeT respectively as suggested in their papers. The \emph{predlift} mode of G-PCCv23 is used following the common test conditions \cite{GPCCconditions}.

In addition, a recent lossless PCAC model 3CAC~\cite{wang2023losslessPCAC} is also compared following their recommended test condition in Sec.~\ref{sec:voxel_trained}, since their source code is not available when conducting this work.


\begin{figure*}[t]
    \centering
    \includegraphics[width=1.0\linewidth]{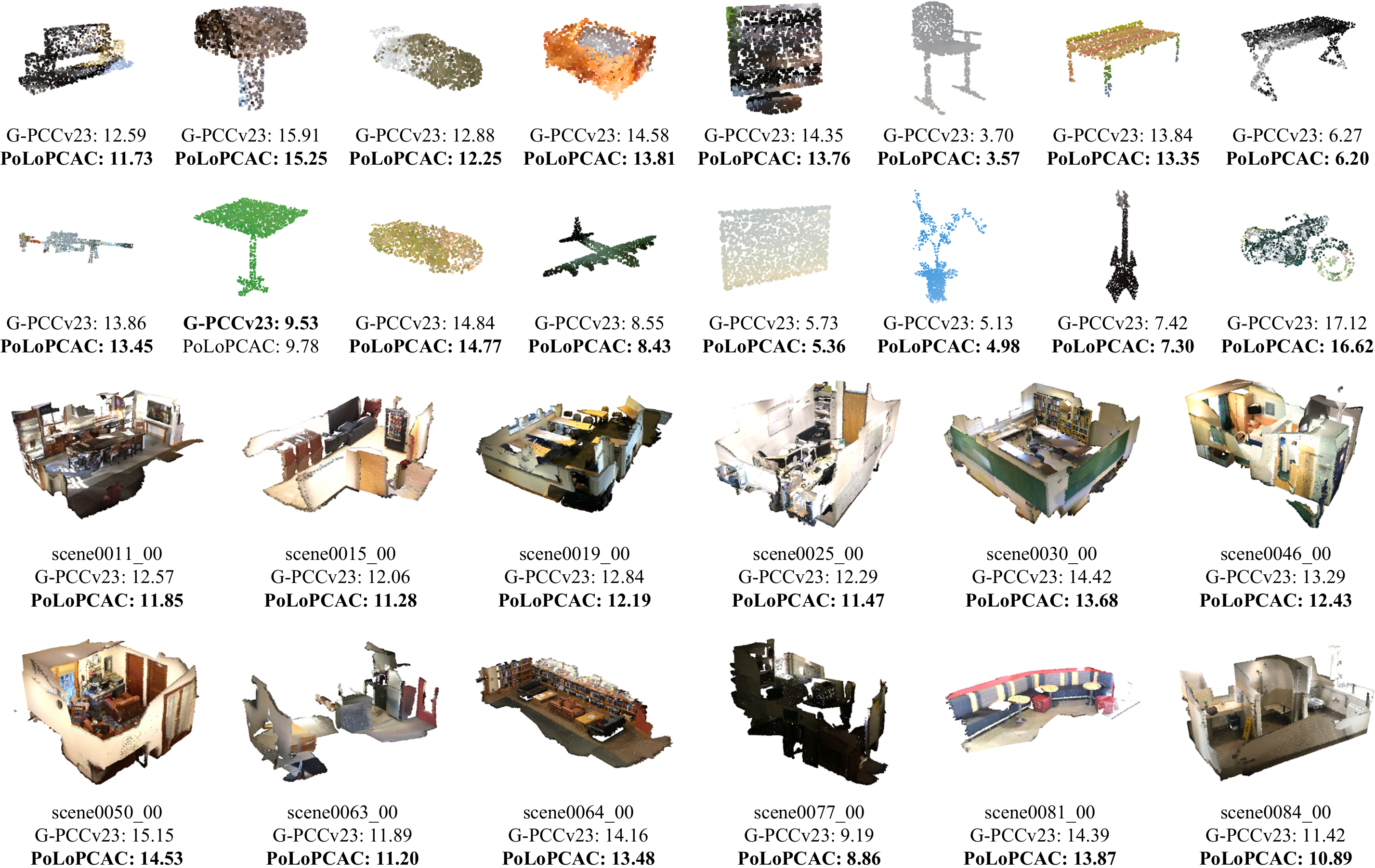}
    \caption{Examples of compression results on 10bit Synthetic 2k-ShapeNet test set (the first two rows) and 12bit ScanNet test set (the third and fourth rows). Bits per point (Bpp) is reported for both G-PCCv23 standard and Synthetic 2k-ShapeNet trained PoLoPCAC. Note that the object point clouds are \emph{randomly} selected and the scene point clouds are with the lowest scene numbers in the official evaluation split, which best ensures the fairness of comparison.}
    \label{fig:visual_obj_scene}
\end{figure*}

\subsection{Quantitative Evaluation}

{
\setlength{\tabcolsep}{0.28em} 
\begin{table}[t]
    \centering
    \footnotesize
    \begin{tabular}{ ccccccc }
    \toprule

        Dataset & Point Cloud & G-PCCv23 & CNeT & MNeT & PoLoPCAC \\
                                                                
    \midrule
        \multirow{2}{*}{ShapeNet}   & 6bit Avg. & \underline{10.85}     & 11.45 & 17.34 & \textbf{10.09} \\
                                    & 10bit Avg. & \underline{10.44}    & 17.72 & 57.66 & \textbf{10.09} \\
    \midrule
        \multirow{2}{*}{ScanNet}    & 10bit Avg. & \underline{13.21}    & 19.61 & 35.34 & \textbf{12.63} \\
                                    & 12bit Avg. & \underline{13.17}    & 23.60 & 58.08 & \textbf{12.63} \\
    \midrule
    \midrule
        \multirow{6}{*}{\shortstack{MVUB\\(vox9)}}  & {Andrew}  & \underline{12.29} & 13.03 & 16.06 & \textbf{12.09} \\
        & {David}   & \textbf{8.00} & \underline{10.60} & 11.23 & \textbf{8.00} \\
        & {Phil}    & 11.98 & \underline{11.95} & 15.69 & \textbf{11.60} \\
        & {Ricardo} & \underline{6.98} & 11.19 & 10.46 & \textbf{6.89} \\
        & {Sarah}   & \underline{5.51} & 8.00 & 9.96 & \textbf{5.21} \\
        & {Avg.} & \underline{8.95} & 10.95 & 12.68 & \textbf{8.76} \\
    \midrule
         \multirow{5}{*}{\shortstack{8iVFB\\(vox10)}}& {Longdress}   & \underline{11.48} & 14.97 & 15.90 & \textbf{11.09}\\
        & {Red\&black}  & \underline{9.18}  & 14.83 & 14.63 & \textbf{8.87} \\
        & {Loot}        & \underline{6.08} & 14.02 & 8.67 & \textbf{5.99} \\
        & {Soldier}     & \textbf{6.74}  & 16.60 & 9.35 & \underline{6.88} \\
        & {Avg.}     & \underline{8.37}  & 15.11 & 12.14 & \textbf{8.21} \\

    \midrule
    \midrule
        \multicolumn{2}{c}{Avg. Time(s/frame)} & Enc/Dec & Enc/Dec & Enc/Dec & Enc/Dec \\
    \midrule
    \midrule
        \multirow{2}{*}{ShapeNet}   & 6bit Avg.     & \underline{0.04}/\underline{0.04} & 2.21/- & 0.27/-  & \textbf{0.03}/\textbf{0.03} \\
                                    & 10bit Avg.    & \underline{0.04}/\underline{0.04} & 2.57/- & 0.28/- & \textbf{0.03}/\textbf{0.03} \\
    \midrule
        \multirow{2}{*}{ScanNet}    & 10bit Avg.   & 1.51/\underline{1.47} & 7.33/- & \underline{1.12}/- & \textbf{0.80}/\textbf{0.84} \\
                                    & 12bit Avg.   & \underline{1.53}/\underline{1.50} & 12.80/- & 2.28/- & \textbf{0.81}/\textbf{0.84} \\
    \midrule
    \midrule
        MVUB & vox9 Avg. & \underline{2.66}/\underline{2.60} & 32.71/- & 3.53/- & \textbf{1.59}/\textbf{1.67} \\
    \midrule
        8iVFB & vox10 Avg. & \underline{8.12}/\underline{7.94} & 63.64/- & 9.92/- & \textbf{6.27}/\textbf{6.85} \\
    \bottomrule

    \end{tabular}
    \caption{ Evaluation of compression performance on \emph{Object\&Scene} and \emph{HumanBodies} point clouds for color attribute. Bpp (Bits per point) is reported in the table. The \textbf{best} and \underline{second-best} results are highlighted in bold and underlined, respectively.}
    \label{tab:result_color_attr}
\end{table}
}

\noindent{\bf Object\&Scene.} Point clouds in \emph{Object\&Scene} need to be rescaled to input into G-PCC, CNeT, and MNeT, in which they are further processed by voxelization operation. Since the scale can affect the sparsity of voxel, we conducted experiments at different scales to present comprehensive results.

As shown in Tab.~\ref{tab:result_color_attr}, our method provides continuous bit-rate reduction over the G-PCCv23 standard. However, CNeT and MNet encounter difficulties in generalizing 6bit ShapeNet-trained models to 10bit test set and ScanNet scenes. It can also be observed that our method achieves a stable bit rate regardless of the scale. Other methods, on the contrary, exhibit fluctuating performances on point clouds of different scales, due to the different inputs produced by voxelization.

\noindent{\bf HumanBodies.} Table~\ref{tab:result_color_attr} also demonstrates the compression performance on dense voxelized point clouds in MVUB and 8iVFB, where our method achieves the lowest bit rate on the majority of sequences, while CNeT and MNeT are suffering from inferior performance.


{
\begin{table}[t]
    \centering
    \small
    \begin{tabular}{ cccccc }
    \toprule

        Category & Point Cloud & G-PCCv23   & PoLoPCAC  \\
    \midrule
        \multirow{2}{*}{LiDAR} & 10bit Avg. & \underline{5.31}    & \textbf{5.28} \\
                                & 12bit Avg. & \underline{5.37}    & \textbf{5.28} \\
    \midrule
        \multicolumn{2}{c}{Avg. Time (s/frame)} & Enc/Dec  & Enc/Dec \\
    \midrule
        \multirow{2}{*}{LiDAR} & 10bit Avg.  & \textbf{0.42}/\textbf{0.42} & \underline{0.47}/\underline{0.46} \\
        & 12bit Avg.  & \underline{0.74}/\underline{0.74} & \textbf{0.47}/\textbf{0.46} \\
    \bottomrule
    \end{tabular}
    \caption{Qualitative results of reflectance compression on SemanticKITTI. Bpp (Bits per point) is reported in the table.}
    \label{tab:result_reflectance}
\end{table}
}

\noindent{\bf LiDAR.} We perform reflectance attribute compression by deploying our color-trained model to LiDAR point clouds. Since the reflection signal only corresponds to a single channel, we repeat the channel three times to adapt to the input of our network, but only code the first channel during arithmetic coding. As shown in Tab.~\ref{tab:result_reflectance}, although our model is trained on the color attributes of small-scale objects, it can still achieve satisfactory compression results on the reflectance attribute of automotive LiDAR scans. Again, compared to G-PCC, our approach demonstrates an excellent \emph{scale-invariant} characteristic.


\subsection{Qualitative Results on Example Point Clouds}

\begin{figure*}[t]
    \centering
    \includegraphics[width=1.0\linewidth]{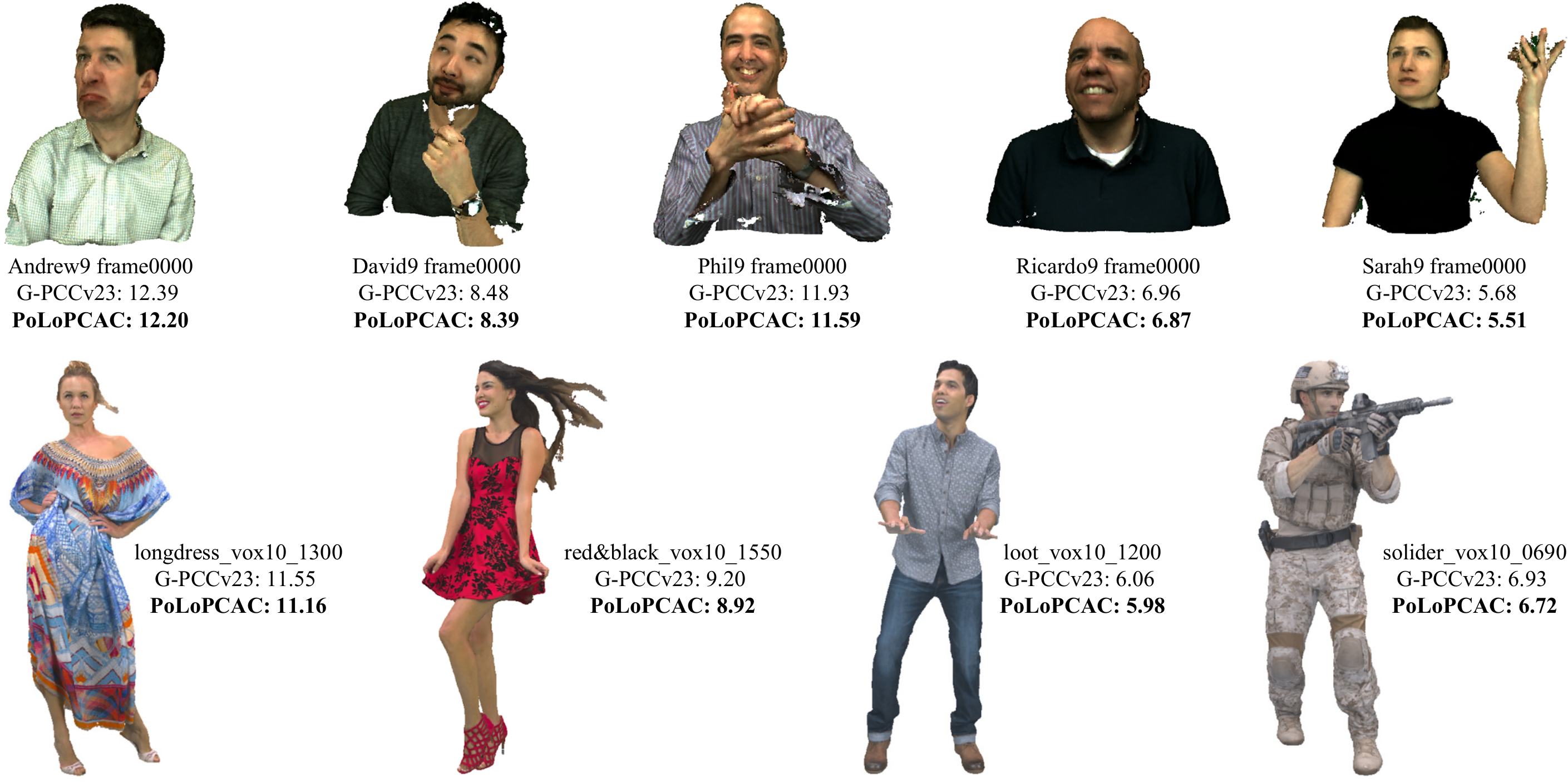}
    \caption{Examples of compression results on human body point clouds of MVUB (the first row) and 8iVFB (the second row). Bits per point (Bpp) is reported for both G-PCCv23 standard and Synthetic 2k-ShapeNet trained PoLoPCAC.}
    \label{fig:visual_human}
\end{figure*}

\begin{figure}[t]
    \centering
    \includegraphics[width=1.0\linewidth]{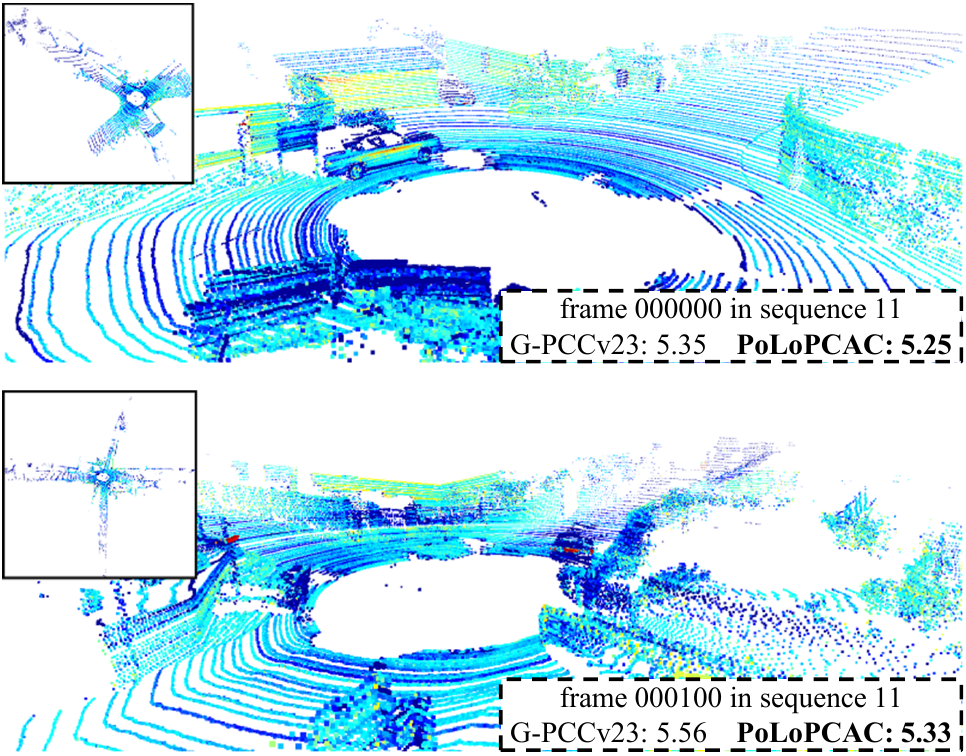}
    \caption{Examples of compression results on SemanticKITTI (12bit).  Bits per point (Bpp) is reported for both G-PCCv23 standard and Synthetic 2k-ShapeNet trained PoLoPCAC.}
    \label{fig:visual_lidar}
\end{figure}

Compression results of example point clouds from different datasets are shown in Fig.~\ref{fig:visual_obj_scene}, Fig.~\ref{fig:visual_human}, and Fig.~\ref{fig:visual_lidar}, where the proposed PoLoPCAC demonstrates superior performance on various test samples once trained on the Synthetic 2k-ShapeNet training set.


\subsection{Voxel Data Training}
\label{sec:voxel_trained}

{
\setlength{\tabcolsep}{0.2em} 
\begin{table}[t]
    \footnotesize
    \centering
    \begin{tabular}{ cccccc }
    \toprule
        Dataset & Point Cloud & G-PCCv23 & \multicolumn{1}{c}{CNeT\textsuperscript{VT}} & \multicolumn{1}{c}{MNeT\textsuperscript{VT}} & \multicolumn{1}{c}{PoLoPCAC\textsuperscript{VT}}  \\
    \midrule
        \multirow{2}{*}{\shortstack{ShapeNet}}      & 6bit Avg.       & \underline{10.85}   & \textbf{10.42} & 22.21 & 13.80 \\ 
                                                    & 10bit Avg.      & \textbf{10.44}   & 22.16 & 68.05 & \underline{13.80} \\
    \midrule
        \multirow{2}{*}{\shortstack{ScanNet}}       & 10bit Avg.      & \textbf{13.21}   & 15.43 & 40.92 & \underline{13.67} \\  
                                                    & 12bit Avg.      & \textbf{13.17}   & 20.00 & 65.50 & \underline{13.67} \\  
    \midrule
    \midrule
        \multirow{3}{*}{\shortstack{MVUB\\(vox9)}}      & Phil      & \underline{11.98} & \textbf{6.61} & 13.89 & 12.79 \\
                                                        & Ricardo   & \underline{6.98}  & \textbf{4.36} & 8.63 & 8.16 \\
                                                        & Avg.   & \underline{9.48} & \textbf{5.49} & 11.26 & 10.48 \\
    \midrule
        \multirow{3}{*}{\shortstack{8iVFB\\(vox10)}}    & Red\&black    & \underline{9.18}  & \textbf{7.72} & 13.67 & 10.91 \\
                                                        & Loot          & \underline{6.08}  & \textbf{4.68} & 8.18 & 8.34 \\
                                                        & Avg.       & \underline{7.63} & \textbf{6.20} & 10.93 & 9.63 \\
    \midrule
    \midrule
    \multicolumn{2}{c}{Avg. Time(s/frame)} & Enc/Dec & Enc/Dec & Enc/Dec & Enc/Dec \\
    \midrule
    \midrule
        \multirow{2}{*}{ShapeNet}   & 6bit Avg. & \textbf{0.04}/\underline{0.04} & 2.19/-  & \underline{0.28}/- & \textbf{0.04}/\textbf{0.03} \\
                                    & 10bit Avg. & \textbf{0.04}/\underline{0.04} & 2.56/- & \underline{0.28}/- & \textbf{0.04}/\textbf{0.03} \\
    \midrule
        \multirow{2}{*}{ScanNet}   & 10bit Avg. & 1.51/\underline{1.47} & 6.04/-  & \underline{1.12}/- & \textbf{0.80}/\textbf{0.84} \\
                                    & 12bit Avg. & \underline{1.53}/\underline{1.50} & 13.42/- & 2.70/- & \textbf{0.81}/\textbf{0.84} \\
    \midrule
    \midrule
        MVUB & vox9 Avg. & 2.79/\underline{2.73} & 19.06/- & \textbf{2.52}/- & \underline{2.67}/\textbf{2.60} \\
    \midrule
        8iVFB & vox10 Avg. & \underline{7.43}/\underline{7.27} & 33.39/- & 8.38/- & \textbf{6.75}/\textbf{6.60} \\
    \bottomrule
    \end{tabular}
    \caption{Comparison of voxel data-trained CNeT\textsuperscript{VT}, MNeT\textsuperscript{VT}, and PoLoPCAC\textsuperscript{VT}.}
    \label{tab:vt_compare}
\end{table}
}

{
\setlength{\tabcolsep}{0.34em} 
\begin{table}[t]
    \small
    \centering
    \begin{tabular}{ ccccc }
    \toprule
        Dataset & Point Cloud & G-PCCv23 & \multicolumn{1}{c}{3CAC\textsuperscript{VT}} & \multicolumn{1}{c}{PoLoPCAC\textsuperscript{VT}}  \\
    \midrule
        \multirow{2}{*}{\shortstack{ScanNet}}       & q5cm Avg.      & 12.79   & \textbf{11.21} & \underline{12.33} \\  
                                                    & q2cm Avg.      & \underline{13.10}   & \textbf{11.86} & 13.36 \\
    \midrule
    \midrule
        \multirow{3}{*}{\shortstack{MVUB\\(vox10)}}     & Phil      & \underline{10.21} & \textbf{6.78} & 11.29 \\
                                                        & Ricardo   & \underline{6.21}  & \textbf{3.59} & 10.43 \\
                                                        & Avg.      & \underline{8.21} & \textbf{5.19} & 10.86 \\
    \midrule
        \multirow{3}{*}{\shortstack{8iVFB\\(vox10)}}    & Red\&black    & \underline{9.18}  & \textbf{8.07} & 11.01 \\
                                                        & Loot          & \underline{6.08}  & \textbf{5.18} & 8.22 \\
                                                        & Avg.          & \underline{7.63}  & \textbf{6.63} & 9.62 \\
    \midrule
        \multirow{3}{*}{\shortstack{Owlii\\(vox11)}}    & Player    & \underline{7.67}  & \textbf{6.78} & 8.61 \\
                                                        & Dancer    & \underline{7.74}  & \textbf{6.80} & 8.48 \\
                                                        & Avg.      & \underline{7.71} & \textbf{6.79} & 8.55 \\
    \midrule
    \midrule
    \multicolumn{2}{c}{Avg. Time(s/frame)} & Enc/Dec & Enc/Dec & Enc/Dec \\
    \midrule
    \midrule
        \multirow{2}{*}{ScanNet}    & q5cm Avg. & \underline{0.25}/\underline{0.25} & -/-  & \textbf{0.17}/\textbf{0.18} \\
                                    & q2cm Avg. & \underline{1.16}/\underline{1.13} & 6.3/6.3  & \textbf{0.67}/\textbf{0.71} \\
    \midrule
    \midrule
        MVUB    & vox10 Avg.  & \underline{15.56}/\underline{14.76} & 27.3/28.3  & \textbf{11.31}/\textbf{12.31} \\
    \midrule
        8iVFB   & vox10 Avg.  & \underline{7.43}/\underline{7.27}   & 15.7/16.0 & \textbf{6.52}/\textbf{6.28} \\
    \midrule
        Owlii   & vox11 Avg.  & \textbf{25.39}/\textbf{24.84} & 56.0/58.1 & \underline{39.98}/\underline{43.03} \\
    \bottomrule
    \end{tabular}
    \caption{Comparison of voxel data-trained 3CAC\textsuperscript{VT} and PoLoPCAC\textsuperscript{VT}. The results of 3CAC\textsuperscript{VT} are directly quoted from their paper~\cite{wang2023losslessPCAC}, which are tested on an RTX3090 GPU.}
    \label{tab:3cac_compare}
\end{table}
}

In the previous sections, we compared the learning-based methods (e.g., CNeT and MNeT) under the the same training set (i.e., Synthetic 2k-ShapeNet) as PoLoPCAC, which has ensured the fairness of comparison. However, since the performance of Neural Network (NN)-based models is widely acknowledged to heavily depend on the quality of the training data~\cite{2021datasetquality}, this section explores varied training conditions to present comprehensive validation.

Specifically, we retrain CNeT, MNeT, and PoLoPCAC following the training data used in CNeT~\cite{nguyen2023CNeT}, which mainly consists 
of voxelized point clouds with relatively dense surfaces (e.g., human bodies). We denote the models trained on voxel data with the superscript VT to differentiate them from the non-voxel 2k-ShapeNet trained models.

As seen from Tab.~\ref{tab:vt_compare}, both CNeT\textsuperscript{VT} and MNeT\textsuperscript{VT} perform better on the human body sequences after trained on the dense voxel point cloud data. This can be attributed to their convolution-based pipelines which fit well on training samples with dense surface but exhibit inadequate capability to learn from sparse data (e.g., 2k-ShapeNet) due to ineffective aggregation caused by fixed-scale convolution kernels.

Similarly, our method also demonstrates a certain degree of dependence on the training data. The performance of PoLoPCAC\textsuperscript{VT} exhibits a slight decrease compared to that of the non-voxel data training results (see Tab.~\ref{tab:result_color_attr}). The quantized voxels, which we believe are insufficient to provide accurate positional encoding within local windows, affect the learning process of the point-based attention mechanism.

Another learning-based method, a.k.a., 3CAC~\cite{wang2023losslessPCAC}, which utilizes cross-scale, cross-group, and cross-color predictions to perform lossless PCAC compression, is also included for comparison in this section. Since their source code is not publicly available, we follow their training/testing conditions and compare with the reported results from their paper. Note that the they also utilize voxel data as the training set.

As observed from Tab.~\ref{tab:3cac_compare}, although PoLoPCAC\textsuperscript{VT} performed relatively inferior under their training/testing condition, we would like to highlight that our runtime remains significantly fast, especially considering that our model is tested on a comparatively low-end platform (e.g., one RTX 2080Ti), and the model size of PoLoPCAC is more than 77$\times$ smaller than that of 3CAC (see Tab.~\ref{tab:modelsize}).



\subsection{Computational Complexity}

We record the encoding and decoding times of different methods, as shown in Tab.~\ref{tab:result_color_attr}, Tab.~\ref{tab:result_reflectance}, Tab.~\ref{tab:vt_compare}, and Tab.~\ref{tab:3cac_compare}. Although our method is prototyped using Python, it still reports shorter coding time than G-PCCv23 on the majority of sequences, even with one RTX 20-series GPU. It should be noted that the decoding times of CNeT and MNeT cannot be determined since their decompression codes are not available. However, as demonstrated in the paper~\cite{nguyen2023CNeT}, the decoding time of CNeT is significantly longer than its encoding time (about 20$\times$ slower).

Table~\ref{tab:modelsize} illustrates the model sizes of compared learning-based methods. As seen, our model reports a significantly small number of parameters, which, along with notably fast encoding and decoding speeds, presents a cost-effective lossless PCAC solution.

\begin{table}[t]
    \small
    \centering
    \begin{tabular}{cccccc}
    \toprule
         Method & MNeT & CNeT & 3CAC & PoLoPCAC \\
    \midrule
         Model Size & 52MB & 960MB & 201MB & \textbf{2.6MB} \\
    \bottomrule
    \end{tabular}
    \caption{Model size of learning-based PCAC methods.}
    \label{tab:modelsize}
\end{table}



\subsection{Ablation Study}
\label{sec:ablation}

\subsubsection{Impact of Granularity}

In Sec.~\ref{sec:prg}, an upper limit $s^{\ast}$ is set to control the maximum number of points within each group, a.k.a., granularity. Tab.~\ref{tab:ablation_on_s} demonstrates the compression efficiencies under different granularities from $2^{10}$ to $2^{16}$. It can be observed that as $s^{\ast}$ decreases, the bit rate of our method gradually decreases and the consumption of GPU memory also declines. However, smaller granularity results in a larger number of groups thus exerting longer coding times. We consider $2^{14}$ as appropriate value on our experimental platform.

{
\begin{table}[t]
    \small
    \centering
    \begin{tabular}{ cccccccc }
    \toprule
        \multirow{2}{*}{Param} & \multirow{2}{*}{\#Groups} & \multirow{2}{*}{Bpp} & \multicolumn{2}{c}{Memory(MB)} & \multicolumn{2}{c}{Time(s)} \\
        \cmidrule(lr){4-5} \cmidrule(lr){6-7}
                                                        & & & Enc & Dec                   & Enc & Dec \\
    \midrule
        $s^{\ast}$=$2^{16}$ & 16 & 11.27  & 3816 & 3795 & \textbf{4.89} & \textbf{5.04} \\
        $s^{\ast}$=$2^{14}$ & 54 & 11.16  & 993 & 972 & \underline{6.42} & \underline{6.82} \\
        $s^{\ast}$=$2^{12}$ & 210 & \underline{11.09}  & \underline{286} & \underline{266} & 14.40 & 15.24 \\
        $s^{\ast}$=$2^{10}$ & 838 & \textbf{11.06}  & \textbf{110} & \textbf{90} & 46.73 & 49.91 \\
    \bottomrule
    \end{tabular}
    \caption{Study on granularity $s^{\ast}$. ``longdress\_vox10\_1300'' in 8iVFB is used for test. \#Groups represents the number of divided groups.}
    \label{tab:ablation_on_s}  
\end{table}
}

\subsubsection{Group-wise Decomposition}

\begin{figure}[b]
    \centering
    \includegraphics[width=1.0\linewidth]{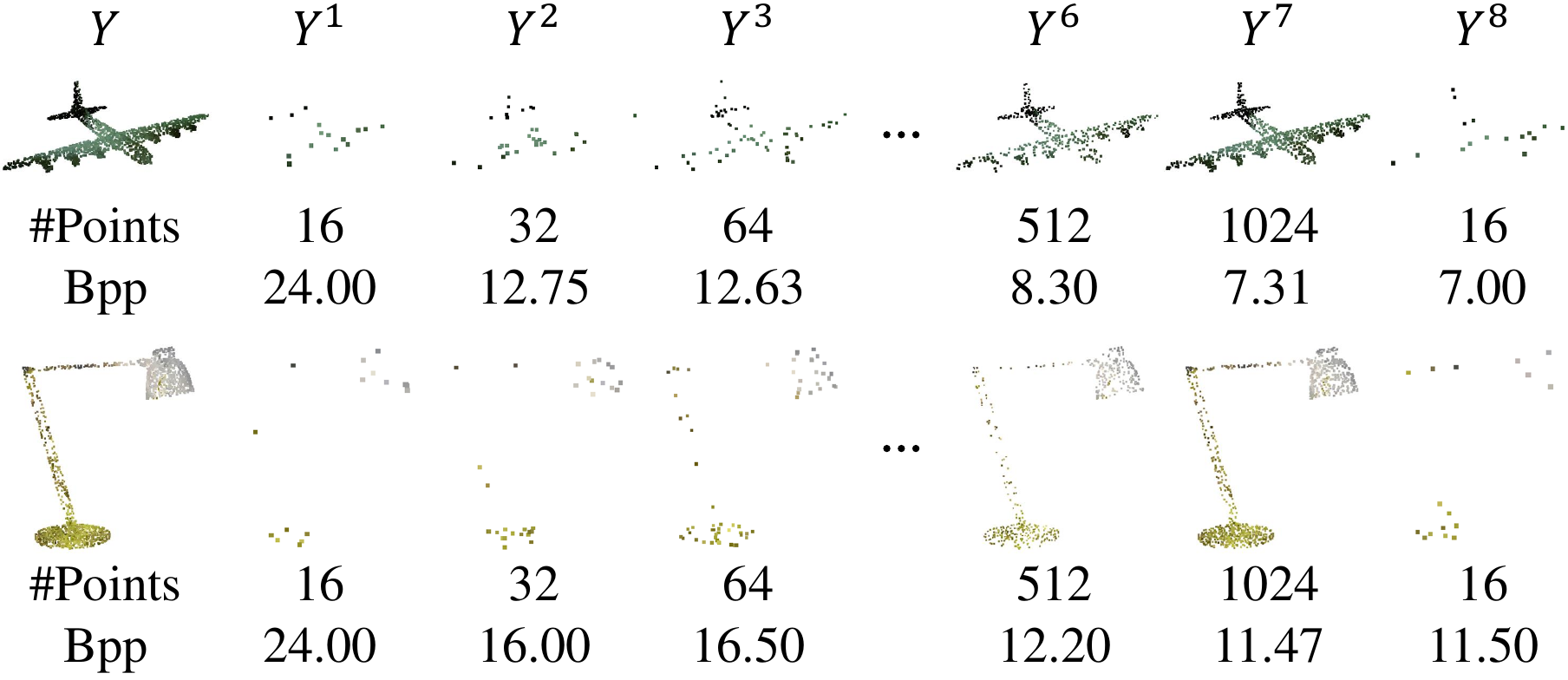}
    \caption{Visualization of grouped results. We calculate the Bpp (Bits per point) of each group for a detailed examination.}
    \label{fig:group_wise_decompo}
\end{figure}

We visualize the groups of example point clouds in Fig.~\ref{fig:group_wise_decompo}. Point clouds are selected from Synthetic 2k-ShapeNet dataset. Since the first group $Y^1$ is coded under uniform distribution, the Bpp of this group is 24. A continuous decrease of Bpp can be observed in the subsequent groups, due to the increasingly abundant contextual information available.

\subsubsection{Spatial Normalization} 
We employ the Spatial Normalization (SN) to normalize each context window. As shown in Eq.~\ref{eq:spatial_norm}, the SN operation can be dissected into two operations: aligning the context window to the target position (i.e., centering), and rescaling of the centered window. We conduct ablation studies following this way and the results are shown in Tab.~\ref{tab:sn_ablation}. It can be seen that the centering operation is crucial for network learning, and thus effectively improves compression performance. The rescaling operation ensures generalizability across different scales and densities.

{
\begin{table}[t]
    \small
    \centering
    \begin{tabular}{ cccccc }
    \toprule
        \multicolumn{2}{c}{Operation} & \multicolumn{2}{c}{ShapeNet}  & \multicolumn{2}{c}{ScanNet}  \\
        \cmidrule(lr){1-2} \cmidrule(lr){3-4} \cmidrule(lr){5-6}
        Centering & Rescaling        & 6bit & 10bit                   & 10bit & 12bit   \\
    \midrule
        \XSolidBrush & \XSolidBrush & 11.34 & 37.36              & 37.17 & 46.85 \\
        \CheckmarkBold & \XSolidBrush & \underline{10.14} & \underline{15.00}                & \underline{12.69} & \underline{13.84} \\
        \CheckmarkBold & \CheckmarkBold & \textbf{10.09} & \textbf{10.09} & \textbf{12.63} & \textbf{12.63} \\
    \bottomrule
    \end{tabular}
    \caption{Study on Spatial Normalization (SN). The average bitrate for each dataset is reported in the table.}
    \label{tab:sn_ablation}
\end{table}
}

\subsubsection{Window Size}

We perform experiments on different sizes of the context window to explore the effect of window size $K$ on compression efficiency. We retrain our network at four different values of $K$ (i.e., 4, 8, 16, and 32) and use two example point clouds for the test.

Table~\ref{tab:ablation_window_size_scannet} shows the compression results on ``scene0011\_00'' point cloud in ScanNet. We can see that the compression efficiency can be affected by a window when it is too large or too small, since the network performs better at $K$=8 and $K$=16 than at $K$=4 and $K$=32. A window that is too small, we assume, may result in insufficient context information, whereas a window that is too large not only increases the computational consumption, but also increases the complexity of the signals within the window.

\begin{table}[t]
    \small
    \centering
    \begin{tabular}{ ccccccc }
    \toprule
        \multirow{2}{*}{Param} & \multirow{2}{*}{Bpp} & \multicolumn{2}{c}{Memory (MB)}      & \multicolumn{2}{c}{Time (s)} \\
        \cmidrule(lr){3-4} \cmidrule(lr){5-6}
                                                       && Enc & Dec                             & Enc & Dec              \\
    \midrule
        $K$=4                  & 12.10                & \textbf{897} & \textbf{891}     & \textbf{1.17}   & \textbf{1.23} \\
        $K$=8                  & \textbf{11.85}       & \underline{963} & \underline{957} & \underline{1.46}   & \underline{1.50}  \\
        $K$=16                 & \textbf{11.85}       & 1801 & 1795                     & 2.15   & 2.20  \\
        $K$=32                 & \underline{11.90}    & 3472 & 3467                     & 3.27   & 3.32  \\
    \bottomrule
    \end{tabular}
    \caption{Study on window size $K$. ``scene0011\_00'' in ScanNet is used for test, which has 237,360 points.}
    \label{tab:ablation_window_size_scannet}
\end{table}

\subsubsection{Attention Mechanism}

\begin{figure*}[t!]
    \centering
    \includegraphics[width=0.95\linewidth]{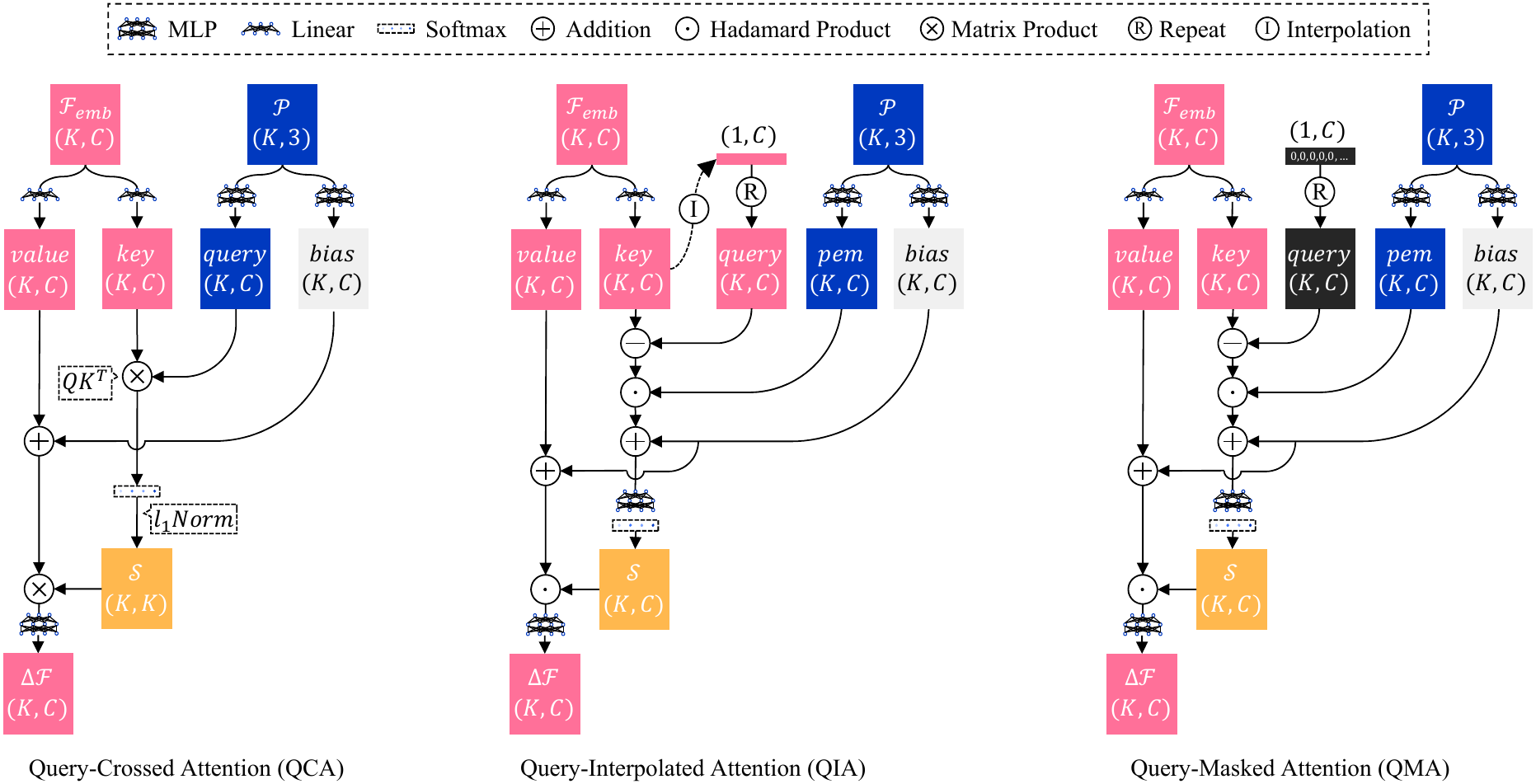}
    \caption{Different mechanisms of attention considered in this work.}
    \label{fig:ablation_attention}
\end{figure*}

In addition to the introduced Query-Masked Attention (QMA), scalar dot-product attention \cite{vaswani2017attention, guo2021pct, li2023exploiting} is another reasonable way of correlating coordinate and attribute information. Here, a Query-Crossed Attention (QCA) is considered as a viable option, following the widely-used Point Cloud Transformer (PCT) \cite{guo2021pct}. Mathematically, we reformulate the operation in Eq.~\ref{eq:attn_calc_score} and Eq.~\ref{eq:attn_calc_fea} following~\cite{guo2021pct}:
\begin{equation}
    \mathcal{S}^{(l)} = N_{L_1} \left( \sigma \left( {\delta}_{query}^{(l)}(\mathcal{P}) \otimes \left( {\psi}_{key}^{(l)}(\mathcal{F}^{(l)}_{emb})\right)^T \right) \right)
\end{equation}
\begin{equation}
    \bigtriangleup \mathcal{F}^{(l)} = {\delta}^{(l)}_{out}\left(\mathcal{S}^{(l)} \otimes \left({\psi}^{(l)}_{value}(\mathcal{F}^{(l)}_{emb})+{\delta}^{(l)}_{bias}(\mathcal{P})\right)\right)
\end{equation}
where ${\delta}^{(l)}$ and ${\psi}^{(l)}$ represents multilayer perceptrons (MLPs) and linear layers, respectively. $\otimes$ refers to the matrix dot product, $\sigma$ refers to the Softmax function, and $N_{L_1}$ refers to the L1-normalization function.

Another intuitive way to conduct subtraction vector attention is to interpolate the \emph{query} based on the \emph{key}. Therefore, we also consider a Query-Interpolated Attention (QIA) and use two different interpolation tactics, i.e., mean value interpolation and distance-weighted interpolation \cite{qi2017pointnet++, sheng2021deepPCAC}, to examine the network efficiency. The interpolation process can be expressed as:
\begin{equation}
    \mathfrak{q} = \sum_{k=1}^{K} \frac{ \omega (x_k) \times  key_k }{\sum_{k=1}^{K} \omega (x_k) }
\end{equation}
where $x_k$ represents the coordinate of the point in the context window after the spatial normalization. $\omega (x_k)$ denotes the weighting factor, which is set to 1 for mean value interpolation, and set to $\frac{1}{{\left \| x_k \right \|}^2_2 }$ for distance-weighted interpolation. Then, the attention can be further conducted following the standard subtraction relation \cite{zhao2021PTv1, wu2022PTv2}:

\begin{equation}
    \mathcal{S}^{(l)} =\sigma \left( {\delta}_{pem}^{(l)}(\mathcal{P}) \odot \left( {\psi}_{key}^{(l)}(\mathcal{F}^{(l)}_{emb}) - \mathfrak{q} \right) + {\delta}_{bias}^{(l)}(\mathcal {P})\right)
\end{equation}

We examine the QCA and QIA models under the same condition as Query-Masked Attention (QMA), and the results are shown in Tab.~\ref{tab:ablation_attn}. As observed, the performance of QCA is much worse than that of QMA and QIA, which demonstrates that the subtraction vector attention is effective in integrating geometric and color signals. We can also see that the QIA performs well on color attribute compression but lacks generalization to LiDAR point clouds. In this case, distance-weighted interpolation is superior than mean value interpolation, but QMA performs better in general. 


{
\begin{table}[t]
    \small
    \centering
    \begin{tabular}{ ccccc }
    \toprule
        \multirow{2}{*}{Module} & ShapeNet      & ScanNet   & KITTI   & Model \\
                                & Bpp           & Bpp       & Bpp     & Size \\
                                                                
    \midrule
        MLP (w/o QMA)                               & 12.46             & 13.82    & 5.77       & 0.4MB \\
        QMA $\rightarrow$ QCA                       & 11.12             & 29.53    & \underline{5.68}       & 2.0MB \\
        QMA $\rightarrow$ QIA\textsuperscript{1}    & 10.18             & 12.70             & 6.78              & 2.6MB \\
        QMA $\rightarrow$ QIA\textsuperscript{2}    & \underline{10.14} & \underline{12.67} & 5.74              & 2.6MB \\
        QMA                                         & \textbf{10.09}    & \textbf{12.63} & \textbf{5.28} & 2.6MB \\
    \bottomrule
    \end{tabular}
    \caption{Performance comparison between different network structures. ``MLP (w/o QMA)'' refers to the removal of our QMA module. ``QMA $\rightarrow$ QCA'' and ``QMA $\rightarrow$ QIA'' represents the use of QCA and QIA, respectively, instead of QMA. \textsuperscript{1}Mean value interpolation. \textsuperscript{2}Distance-weighted interpolation.}
    \label{tab:ablation_attn}
\end{table}
}

\subsubsection{Time Consumption Dissection} 

We dissect our method to examine the time consumption of each individual module, as shown in Tab.~\ref{tab:time_decompose}. ``scene0011\_00'' in ScanNet and ``longdress\_vox10\_1300'' in 8iVFB are used for test, abbreviated as ``scene'' and ``human'' for brevity, respectively. It can be seen that our grouping strategy is extremely fast, which only requires 0.02s to process a large-scale human-body frame. We believe our running time can be significantly reduced by further optimizing our method on C/C++ platforms.


{
\begin{table}[t]
    \small
    \centering
    \begin{tabular}{cccccc}
    \toprule
         Frame (\#Points) & PRG & $K$NN & Net & AE & Total  \\
    \midrule
         scene (237,360) & 0.01s & 0.18s & 0.94s & 0.34s & 1.47s  \\
         human (857,966) & 0.02s & 2.59s & 2.70s & 1.11s & 6.42s  \\
    \bottomrule
    \end{tabular}
    \caption{Time consumption of each component during encoding. PRG refers to our progressive random grouping strategy, $K$NN refers to the $K$-nearest neighbor operation we use to build context windows, Net refers to the neural network inferring, and AE refers to the arithmetic encoding.}
    \label{tab:time_decompose}
\end{table}
}


\section{Conclusion}
\label{sec:conclusion}


We present an efficient and generic point-based method for lossless PCAC. Leveraging on a point model that utilizes the attention mechanism, we compress the attributes in an inter-group autoregressive and intra-group parallel way, which achieves excellent generalizability, affordable complexity, and considerably small model size. Future work may include the extension of the proposed generic lossless compression solution  to the lossy compression domain for point cloud attribute.

\ifCLASSOPTIONcaptionsoff
  \newpage
\fi



\bibliographystyle{IEEEtran}
\bibliography{IEEEabrv,ref.bib}
%



%









\begin{IEEEbiography}[{\includegraphics[width=1in,height=1.25in,clip,keepaspectratio]{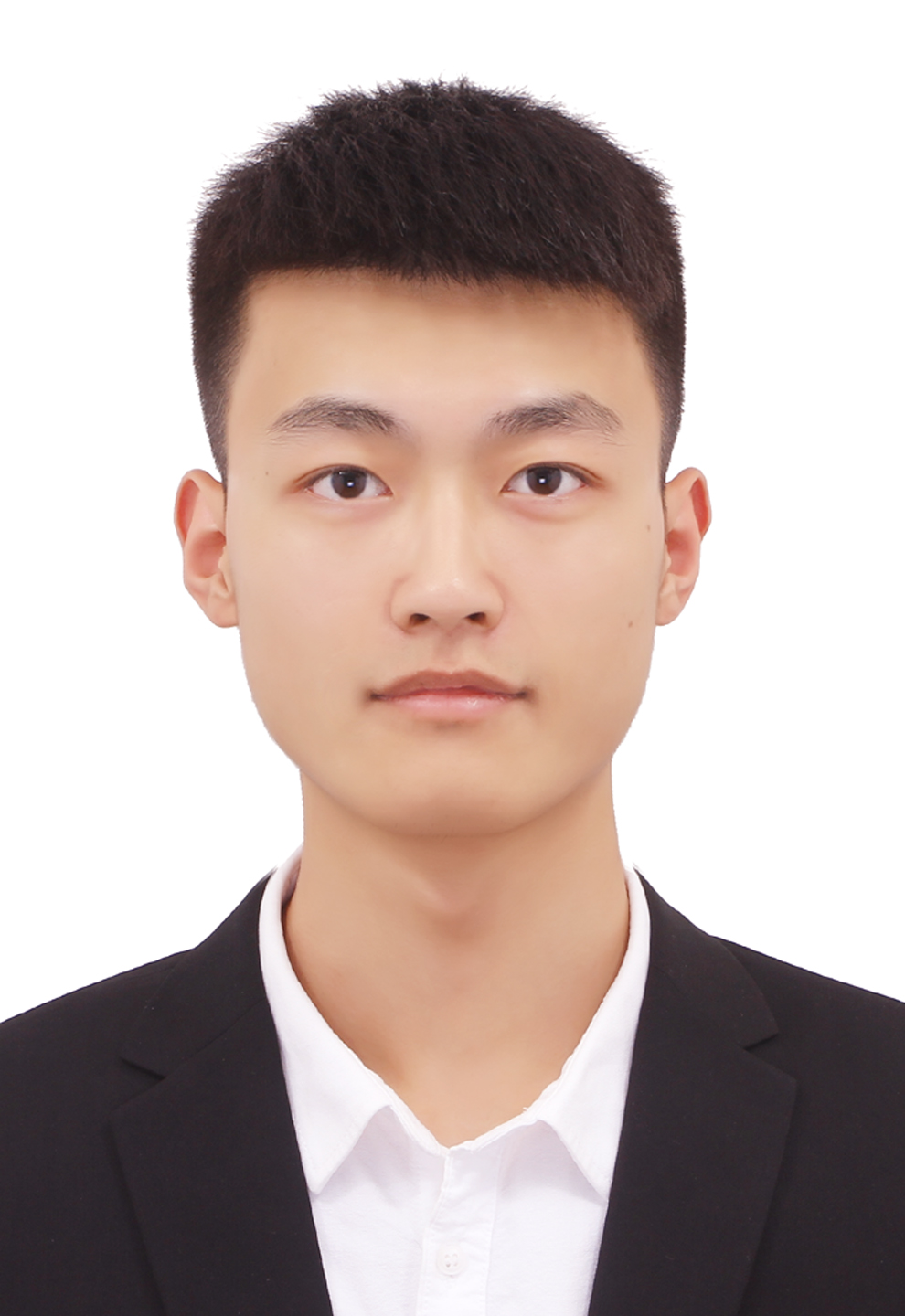}}]{Kang You} received the B.S. degree in Computer Science and Technology from Wenzhou University, Wenzhou, China, in 2019. He received the M.S. degree in Computer Science and Technology from Nanjing University of Aeronautics and Astronautics, Nanjing, China, in 2023. After that, he has been a Research Assistant at Nanjing University of Aeronautics and Astronautics, Nanjing, China. His current research interests include deep learning and point cloud compression. 
\end{IEEEbiography}

\begin{IEEEbiography}[{\includegraphics[width=1in,height=1.25in,clip,keepaspectratio]{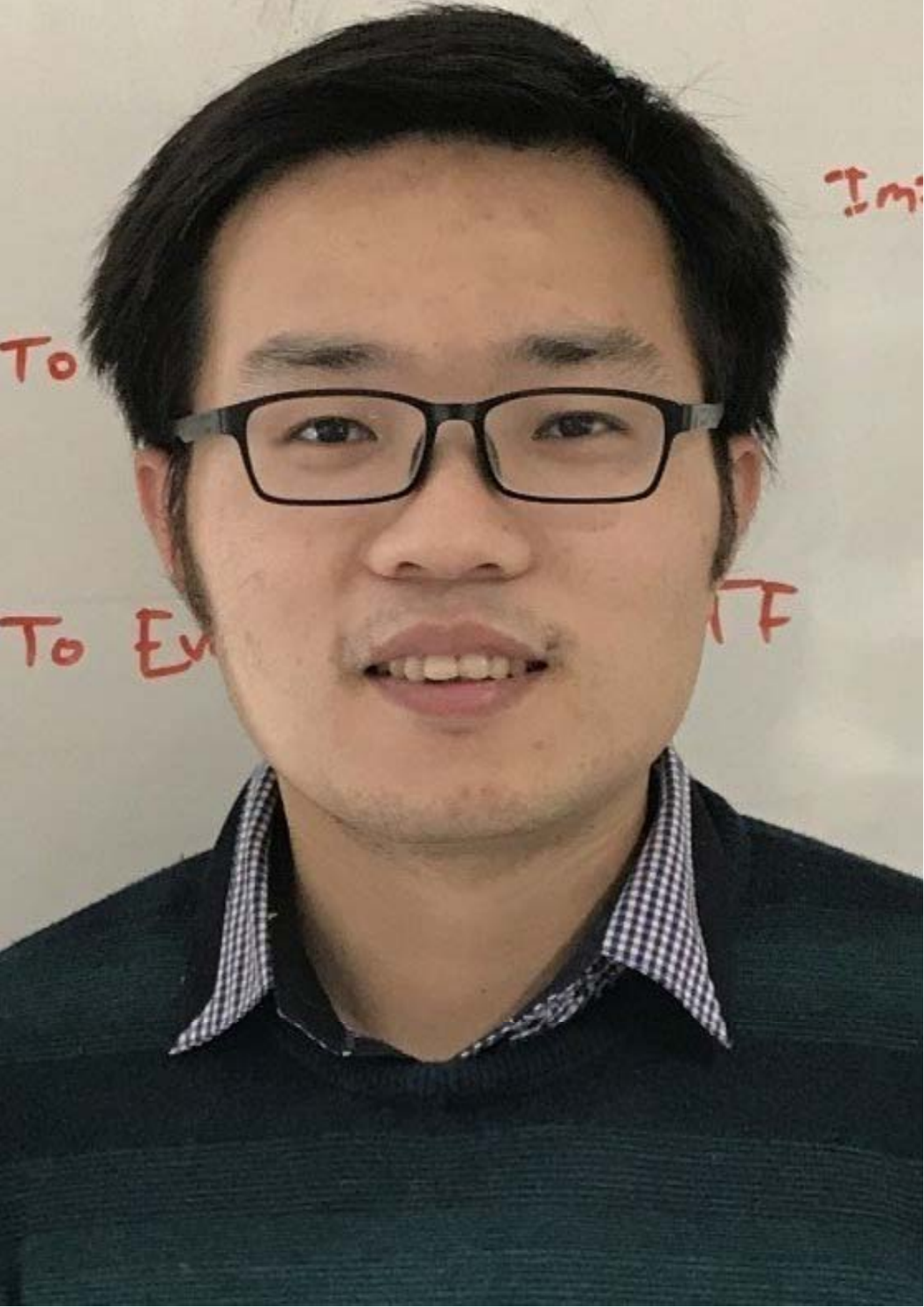}}]{Pan Gao} (M'16) 	received  the Ph.D. degree in electronic engineering from University of Southern Queensland (USQ), Toowoomba, Australia, in 2017. Since  2016, he has been  with the College of Computer Science and Technology, Nanjing University of Aeronautics and Astronautics, Nanjing, China, where he is currently an Associate Professor. From 2018 to 2019, he was a Postdoctoral Research Fellow at Trinity College Dublin, Dublin, Ireland.   His research interests include point cloud compression and understanding, computer vision, and deep learning.
\end{IEEEbiography}

\begin{IEEEbiography}[{\includegraphics[width=1in,height=1.25in,clip,keepaspectratio]{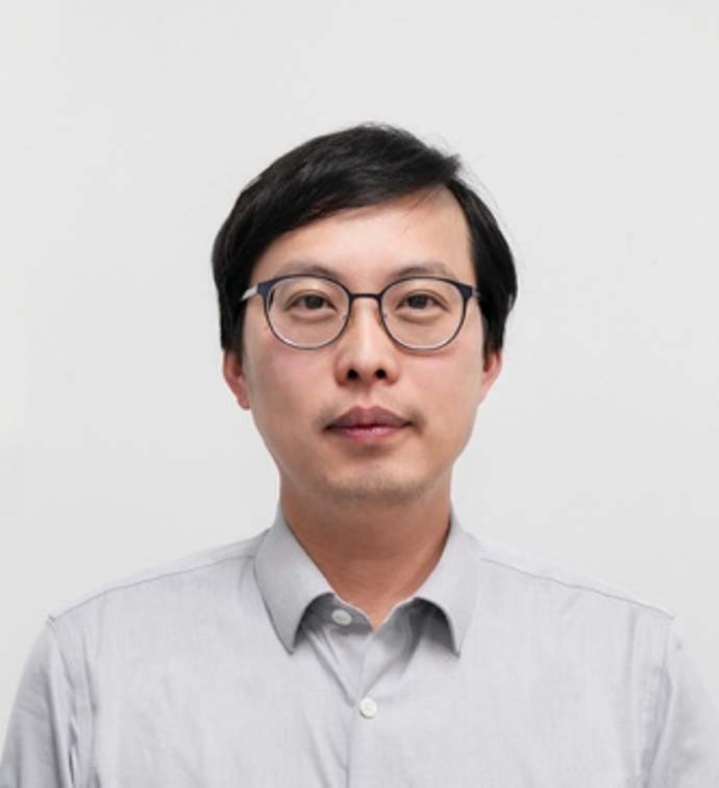}}]{Zhan Ma} (SM'19) is now on the faculty of Electronic Science and Engineering School, Nanjing University, Nanjing, Jiangsu, 210093, China. 
He received the B.S. and M.S. degrees from the Huazhong University of Science and Technology, Wuhan, China, in 2004 and 2006 respectively, and the Ph.D. degree from the New York University, New York, in 2011. From 2011 to 2014, he has been with Samsung Research America, Dallas TX, and  Futurewei Technologies, Inc., Santa Clara, CA, respectively. His research focuses on the learning-based video coding, and smart cameras. He is a co-recipient of the 2018 PCM Best Paper Finalist, 2019 IEEE Broadcast Technology Society Best Paper Award, 2020 IEEE MMSP Image Compression Grand Challenge Best Performing Solution, and 2023 IEEE WACV Best Paper (Algorithms). 
\end{IEEEbiography}

\appendices

\section{Technical Details for Generating Synthetic 2k-ShapeNet Dataset}
\label{sec:supp_details_for_shapenet}

\begin{figure}[h]
    \centering
    \includegraphics[width=1.0\linewidth]{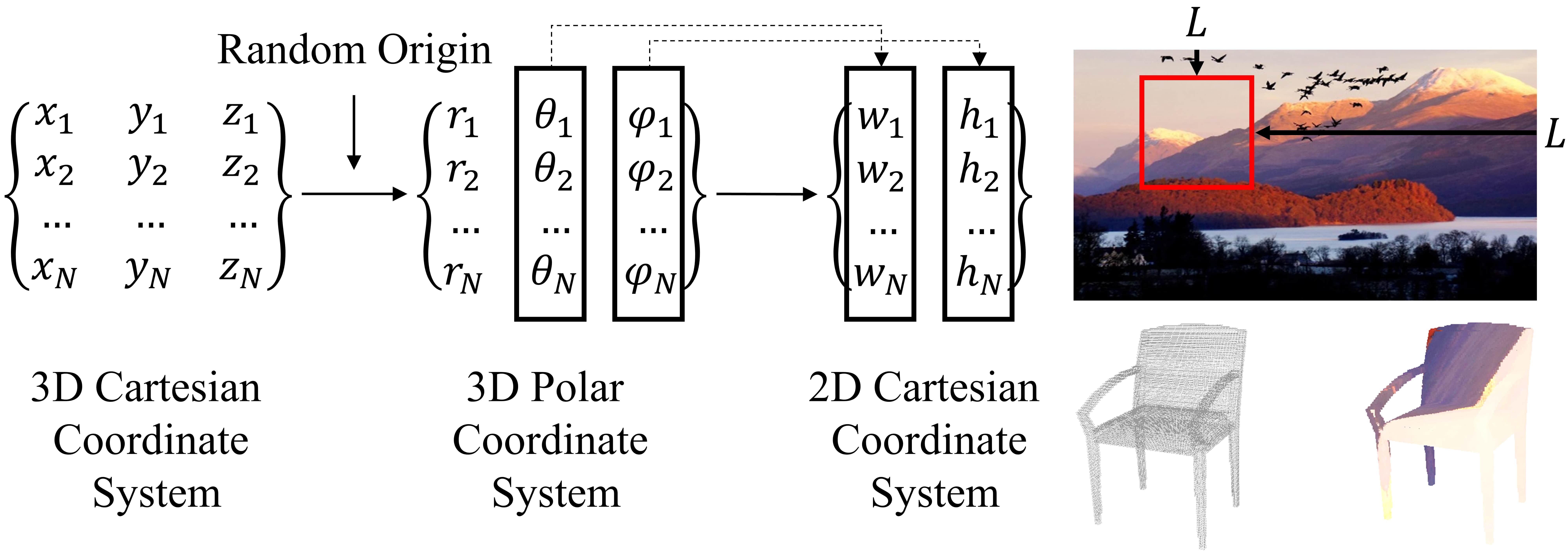}
    \caption{Coloring process for point clouds. We map the two polar angles $({\theta}_i, {\varphi}_i)$ to a square picture grid.}
    \label{fig:supp_colorize_shapenet}
\end{figure}

In brief, we first sample 2048 (2k) points randomly from the mesh of ShapeNet \cite{chang2015shapenet} to form the colorless point cloud, and then each point cloud is colorized by a Polar-to-Euclidean coordinate transform. This process is shown in Fig.~\ref{fig:supp_colorize_shapenet}.

Specifically, a point cloud is denoted as $\left\{ p_1, p_2, \dots, p_N \right\}$, where each point $p_i$ corresponds to a position within 3D Cartesian coordinate system, i.e., $p_i=(x_i, y_i, z_i)$. We first convert the 3D Cartesian coordinate system to a Polar coordinate system, i.e., each point $p_i$ is transformed into $(r_i, {\theta}_i, {\varphi}_i)$, where $r_i$ represents the radius and ${\theta}_i$ and ${\varphi}_i$ represent the polar angles. Then, we map the two angles $({\theta}_i, {\varphi}_i)$ to a squared area with side length $L$, i.e.,
\begin{equation}
    w_i = \nint*{ \frac{{\theta}_{i}-{\theta}_{min}}{{\theta}_{max} - {\theta}_{min}} \times L }, \quad h_i = \nint*{ \frac{{\varphi}_{i}-{\varphi}_{min}}{{\varphi}_{max} - {\varphi}_{min}} \times L }
\end{equation}
where $\theta_{min}$ and $\theta_{max}$ represent the minimum and maximum values of $\theta$ of point cloud, respectively. $\varphi_{min}$ and $\varphi_{max}$ are similarly defined. $\nint{}$ refers to the round operation.

To ensure the diversity of coloring, we randomly select an image from the PCCD dataset \cite{chang2017PCCD} for each point cloud, and then an image block with the size of $L \times L$ is cropped as the mapping source. Let $I_L$ represent the selected image block, and then the color attribute $(r_i, g_i, b_i)$ for point $p_i$ is determined by mapping the color from the corresponding position in $I_L$:
\begin{equation}
    (r_i, g_i, b_i) = {I}_{L}[w_i, h_i]
\end{equation}
where $\left[ \right]$ refers to the mapping operation by utilizing $w_i$ and $h_i$ as the indexes. We set $L$ as a random number from 128 to 512.

In addition, during the Polar coordinate transformation, we randomly select a distant point as the origin of the polar coordinate system, which can reduce the visual distortion of the coloring. Specifically, for simplicity, we randomly select a point as the origin from 8 candidate points formed by the combinations of  $(\pm 1024, \pm 1024, \pm 1024)$, which are equivalent to the 8 vertices of a cube with a side length of 2048, centered at (0,0,0).

\end{document}